\PassOptionsToPackage{utf8}{inputenc}
\documentclass{bioinfo}
\usepackage{amsmath}
\copyrightyear{2015} \pubyear{2015}
\access{Advance Access Publication Date: Day Month Year}
\appnotes{Manuscript Category}
\usepackage{multirow}
\usepackage{algorithm}
\usepackage{algorithmic}
\usepackage{booktabs}
\usepackage{longtable}
\usepackage{graphicx}
\usepackage{multirow}
\usepackage{algorithm}
\usepackage{algorithmic}
\usepackage{booktabs}
\usepackage{supertabular}
\usepackage{longtable}
\usepackage{amsmath}
\usepackage[toc,page]{appendix}

\begin{document}
\firstpage{1}

\subtitle{Data and text mining}

\title[short Title]{Partial Annotation Learning for Biomedical Entity Recognition}
\author[Sample \textit{et~al}.]{Liangping Ding\,$^{\text{\sfb 1,2,3}*}$, Giovanni Colavizza\,$^{\text{\sfb 3}}$, Zhixiong Zhang\,$^{\text{\sfb 1,2}*}$}
\address{$^{\text{\sf 1}}$Department of Information Resources Management, University of Chinese Academy of Sciences, Beijing, 100049, China. \\$^{\text{\sf 2}}$National Science Library, Chinese Academy of Sciences, Beijing, 100190, China. \\$^{\text{\sf 3}}$ Institute for Logic, Language and Computation (ILLC), University of Amsterdam, Amsterdam, 1098 XG, The Netherlands.}

\corresp{$^\ast$To whom correspondence should be addressed.}

\history{Received on XXXXX; revised on XXXXX; accepted on XXXXX}

\editor{Associate Editor: XXXXXXX}
\abstract{\textbf{Motivation:} Named Entity Recognition (NER) is a key task to support biomedical research. In Biomedical Named Entity Recognition (BioNER), obtaining high-quality expert annotated data is laborious and expensive, leading to the development of automatic approaches such as distant supervision. However, manually and automatically generated data often suffer from the \textit{unlabeled entity problem}, whereby many entity annotations are missing, degrading the performance of full annotation NER models. \\
\textbf{Results:} To address this problem, we systematically study the effectiveness of partial annotation learning methods for biomedical entity recognition over different simulated scenarios of missing entity annotations. Furthermore, we propose a TS-PubMedBERT-Partial-CRF partial annotation learning model. We harmonize 15 biomedical NER corpora encompassing five entity types to serve as a gold standard and compare against two commonly used partial annotation learning models, BiLSTM-Partial-CRF and EER-PubMedBERT, and the state-of-the-art full annotation learning BioNER model PubMedBERT  tagger. Results show that partial annotation learning-based methods can effectively learn from biomedical corpora with missing entity annotations. Our proposed model outperforms alternatives and, specifically, the PubMedBERT tagger by 38\% in F1-score under high missing entity rates. The recall of entity mentions in our model is also competitive with the upper bound on the fully annotated dataset.\\
\textbf{Availability:} https://gitee.com/Liangping\_Ding/partial-annotation-learning \\
\textbf{Contact:} \href{dingliangping@mail.las.ac.cn}{dingliangping@mail.las.ac.cn}; \href{zhangzhx@mail.las.ac.cn}{zhangzhx@mail.las.ac.cn}\\
\textbf{Supplementary information:} Supplementary data are available at \textit{Bioinformatics}
online.}
\maketitle

\makeatletter
\let\oldlt\longtable
\let\endoldlt\endlongtable
\def\longtable{\@ifnextchar[\longtable@i \longtable@ii}
\def\longtable@i[#1]{\begin{figure}[t]
\onecolumn
\begin{minipage}{0.5\textwidth}
\oldlt[#1]
}
\def\longtable@ii{\begin{figure}[t]
\onecolumn
\begin{minipage}{0.5\textwidth}
\oldlt
}
\def\endlongtable{\endoldlt
\end{minipage}
\twocolumn
\end{figure}}
\makeatother

\section{Introduction}



Biomedical Named Entity Recognition (BioNER) is a specific sub-task of Named Entity Recognition (NER) that aims to recognize and classify named entities in the biomedical domain. The purpose of BioNER is to automate the process of extracting information from vast amounts of biomedical texts, which plays a crucial role in both relation extraction \citep{cokol2005emergent} and knowledge base completion \citep{szklarczyk2016string}. By accurately identifying and classifying named entities such as genes, diseases, and drugs, BioNER allows for the discovery of new biological relationships between biomedical entities, supporting advances in the field of biomedicine.

For fully annotated NER datasets, this problem has been basically solved by fine-tuning pre-trained language models \citep{devlin2018bert,liu2019roberta}. While in the biomedical domain, due to privacy and ethical concerns \citep{zhang2022biomedical}, the lack of fully annotated datasets is still a common issue for BioNER. This limits the size and diversity of available data, since obtaining high-quality annotations at scale is expensive and labor-intensive. To reduce the reliance on expert annotations, distant supervision \citep{liang2020bond} and exploratory expert \citep{effland2021partially} approaches have been proposed, leading to partially annotated datasets with high precision but low recall for entity spans. Specifically, datasets suffer from the unlabeled entity problem \citep{li2021empirical}, where large amounts of entity annotations are missing, as exemplified by the entity ``SARS-CoV-2'' in Fig. 1. Directly assuming missing labels to be non-entities (O tags) may degrade the performance of NER models.

The unlabeled entity problem in NER has garnered attention and prior work can be mainly divided into two main directions. The first direction aims to design model architectures to alleviate the effects of false negatives in the training dataset, which is generated by labeling all missing labels as non-entities \citep{liang2020bond}. This is accomplished through a model architecture that identifies false negatives and reduces their impact on model performance. The second direction involves using Partial Annotation Learning (PAL), which considers the incompletely labeled data set as a partially annotated data set and directly models missing labels \citep{jie2019better,mayhew2019named}. In this approach, missing labels are treated as latent variables, such as for the Partial Conditional Random Field (Partial CRF) model \citep{bellare2007learning}. All possible label paths are then deduced, and the marginal probabilities are calculated at the missing position, with the parameter estimation methods such as the Expectation Maximization algorithm as in \cite{tsuboi2008training}, utilized to maximize the log-likelihood and estimate model parameters.

Partial Annotation Learning-based methods have been shown to alleviate the unlabeled entity problem effectively in previous studies. With only 1,000 biased and incomplete annotations (less than 10\% of the original annotations for the datasets), a partial annotation learning model still achieves an F1 score of 71.7\% on average \citep{effland2021partially}. While most studies have focused on evaluating the effectiveness of partial annotation learning on traditional NER benchmark datasets such as CoNLL2003 \citep{tjong-kim-sang-de-meulder-2003-introduction}, dealing with the most common named entity types like person, location, and organization. The effectiveness of partial annotation learning methods on the more challenging BioNER benchmark datasets has not been assessed. Furthermore, to the best of our knowledge, no single study exists which comprehensively evaluates the validity of partial annotation learning and conducts an in-depth assessment of the impact of missing entity ratio and annotation scenario on the model performance. Although \cite{effland2021partially} compared the traditional NER models to the partial annotation learning models under several annotation budgets, they set the  number of entity annotations M $\in$ {100 (0.4\%), 500 (2.1\%), 1K (4.3\%), 5K (21.3\%), 10K (42.6\%)}, making it hard to appreciate the performance under degrading conditions systematically.

As we mentioned before, partial annotation learning is a promising approach that allows for the utilization of all possible label paths to train a model. Nevertheless, when a dataset contains abundant missing labels, this can lead to computational costs. To this end, we propose a partial annotation learning-based model architecture for BioNER called TS-PubMedBERT-Partial-CRF, which leverages the advantage of partial annotation learning and uses confidence estimation to iteratively decrease the number of latent variables for parameter estimation in the Partial CRF model. The backbone model architecture is the Partial CRF model built on top of the biomedical pre-trained language model PubMedBERT \citep{gu2021domain}, which is integrated into a  Teacher-Student self-training framework \citep{liang2020bond} with a confidence estimation module to improve model's tolerance to noise. 

Extensive experiments are conducted to evaluate the efficacy of our proposed model and verify the effectiveness of partial annotation learning models to alleviate the unlabeled entity problem for biomedical NER under various settings. Our proposed model is compared with the state-of-the-art Biomedical NER tagger PubMedBERT \citep{gu2021domain}, as well as two partial annotation learning models, BiLSTM-Partial-CRF \citep{jie2019better}, and EER-PubMedBERT \citep{effland2021partially} across 5 biomedical entity types. Our experimental results confirm that even a state-of-the-art full annotation learning model still suffers from the unlabeled entity problem when the number of missing entity annotations increases. Instead, partial annotation learning-based methods can effectively capture missing entity annotations in the dataset, achieving promising results even with 90\% of entity annotations missing. Further, our model performs as well as the state-of-the-art partial annotation learning model from \cite{effland2021partially} across the studied missing entity ratio and the annotation scenario, and performs better under higher missing rates and a more realistic annotation scenario.
 
\begin{figure}
  \centering
  \includegraphics[width=.5\textwidth]{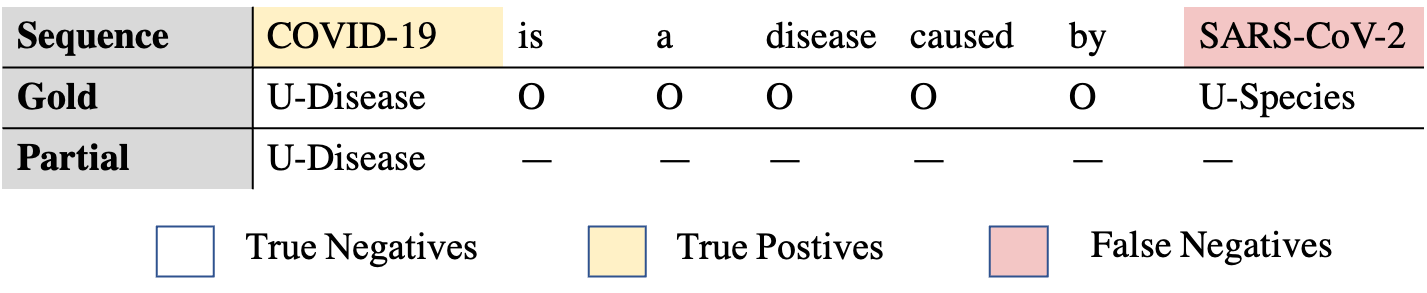}
  \caption{Example illustrating the unlabeled entity problem in NER. The ``Sequence'' row shows the token sequence of the input text, and the ``Gold'' row reflects the golden truth label path under the BILOU encoding scheme, and the ``Partial'' row shows the partially annotated label path, in which the symbol ‘—’ represents the unknown label.}
\end{figure}

\begin{figure*}
  \centering
  \includegraphics[width=\textwidth]{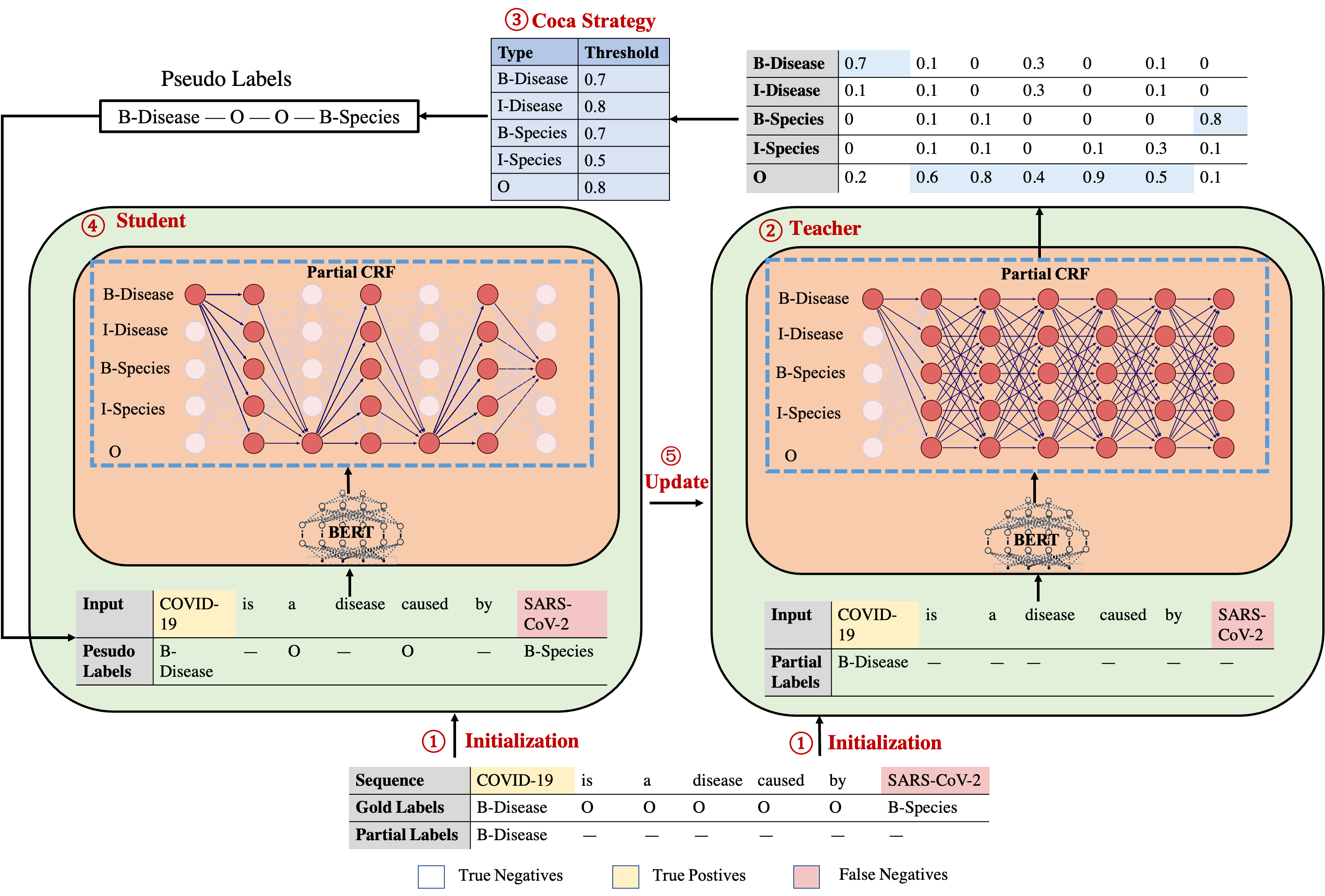}
  \caption{TS-PubMedBERT-Partial-CRF Model Architecture}
\end{figure*}

\vspace{-0.8cm}
\section{Related Work}

Named Entity Recognition (NER) task is typically defined as a sequence labeling task in which tokens in a sequence are annotated using a tagging scheme such as BIO \citep{ramshaw-marcus-1995-text} or BILOU \citep{ratinov2009design}. A conditional Random Field (CRF) model, which captures dependencies between labels, is frequently used as an NER tagger. However, traditional CRF models have limitations in directly modeling missing labels. 

\cite{bellare2007learning} extended the conventional CRF and introduced the M-CRF model to directly learn from incomplete annotations, which is the first application of the partial annotation learning method to the sequence labeling task. To deal with missing labels in an automatic metadata extraction task, they proposed a novel training objective for a CRF model that treated missing labels as latent variables, allowing for partial annotation learning by calculating marginal probabilities over all possible label paths. The Expectation Maximization algorithm \citep{dempster1977maximum} was then utilized to maximize the log-likelihood, and the model parameters were estimated accordingly. 

Partial annotation learning has been applied to many NLP tasks, including part-of-speech tagging \citep{tsuboi2008training}, word segmentation \citep{yang-vozila-2014-semi}, lexical disambiguation \citep{hovy-hovy-2012-exploiting}, to tackle the problem of incomplete annotation in the annotated corpora. In NER, partial annotation learning has been shown to significantly improve the performance of distantly supervised NER models, when compared to simply labeling all missing labels as non-entities to train a supervised NER model, or to using dictionaries to match entities \citep{cao2019low,carlson2009learning,yang2018distantly}. 

While the effectiveness of partial annotation learning models were mainly  evaluated  on common entity types, few works applied partial annotation learning methods on domain-specific NER tasks. \cite{jie2019better} introduced a method to train a BiLSTM-Partial-CRF model with incomplete annotations, assigning high probability mass to the most probable labeling sequence that matches the available partial annotations. They dropped 50\% of the entity annotations in the CoNLL 2003 English dataset \citep{tjong-kim-sang-de-meulder-2003-introduction} and the CoNLL 2002 Spanish NER dataset \citep{tjong-kim-sang-2002-introduction} to simulate the incomplete annotation scenario and evaluate model's performance.  Even though \cite{greenberg-etal-2018-marginal} constructed a partial annotation learning model for BioNER, and achieved promising performance with incomplete annotations compared to the traditional CRF model, they used the BiLSTM-Partial-CRF model architecture, which seems to be outdated for today’s standards as models based on the Transformer architecture \citep{vaswani2017attention}, specifically the pre-trained BERT model.\citep{devlin2018bert}. 

\vspace{-0.8cm}
\section{Materials and methods}

In this section, we provide a technical explanation of the TS-PubMedBERT-Partial-CRF approach and discuss the competitor systems to compare against. Additionally, we outline the details on how to construct synthetic partially annotated datasets from the golden standard corpora using various entity annotation removal algorithms.

\subsection{Task Formulation}
In this work, we formulate the BioNER task as a sequence labeling task, where given a sequence of tokens $\boldsymbol{X}=[x_{1},...x_{i},...,x_{n}]$, the goal is to predict a corresponding label sequence $\boldsymbol{Y}=[y_{1},...y_{i},...,y_{n}], s.t. \;y_{i} \in \mathcal{Y}$  that encodes the named entities, where $\mathcal{Y}$ represents the label set and $n$ is the length of the sequence. The fully annotated NER dataset with $K$ samples can be regarded as a set of pairs of token sequence and label sequence:
\begin{equation}
    \boldsymbol{D}=\left\{\left(\boldsymbol{X}^{(k)},\boldsymbol{Y}^{(k)}\right)\right\}_{k=1}^K
\end{equation}
where $(\boldsymbol{X}^{(k)},\boldsymbol{Y}^{(k)})$ is the $k$-th instance from dataset $\boldsymbol{D}$.

For partially annotated dataset suffering from the unlabeled entity problem, the label sequence is incomplete with unknown labels. Instead of converting all missing labels in the partially annotated dataset to non-entity labels \citep{jie2019better,mayhew2019named}, we mark them as potential entities with a special ``unknown'' type. Given a token sequence $\boldsymbol{X}$, the partial label sequence is a set of observed (label, position) pairs, defined as $\boldsymbol{Y_p}\subset\{\left(y_i,i\right)\mid y_i\in \mathcal{Y},1\le i\le n\}$. And we can derive a collection of all possible complete label sequences  for $\boldsymbol{X}$ that are compatible with $\boldsymbol{Y_p}$, denoted as $\boldsymbol{C}{(\boldsymbol{Y_p})}$. For example, in Fig. 1, there is only one observed label at position 1, so the partial label sequence is $\boldsymbol{Y_p}=\{(\mathrm{U-Disease},1)\}$. For the 6 missing label positions, we can derive a possible label sequence collection $\boldsymbol{C}{(\boldsymbol{Y_p})}$, whose size is $|\mathcal{Y}|^6$. 
Under this formulation, a partially annotated dataset will be given as:
\begin{equation}
    \boldsymbol{D_p}=\left\{\left(\boldsymbol{X}^{(k)},\boldsymbol{Y_p}^{(k)}\right)\right\}_{k=1}^K
\end{equation}
\subsection{TS-PubMedBERT-Partial-CRF model architecture}

TS-PubMedBERT-Partial-CRF model combines insights from the BOND-Coca model \citep{ding2022distantly} and the EER-BERT model \citep{effland2021partially}, by combining a Teacher-Student self-training framework with partial annotation learning to alleviate computation costs. The backbone architecture of the TS-PubMedBERT-Partial-CRF model is PubMedBERT-Partial-CRF, where the pre-trained language model PubMedBERT plays the role of the encoder to generate the contextual language representation and the partial CRF model takes care of the sequential label sequences and models missing labels as latent variables. The PubMedBERT-Partial-CRF model is then integrated into a  Teacher-Student self-training framework with the Coca strategy (Category-Oriented Confidence Calibration)\citep{ding2022distantly} as a confidence estimation method to update the dataset with high-confidence labels, forming the architecture of the TS-PubMedBERT-Partial-CRF model. 

As illustrated in Fig. 2, training a TS-PubMedBERT-Partial-CRF model can be regarded as a two-step process, that is initialization and  Teacher-Student self-training. In the first step, to mitigate the effect of the unlabeled entity problem, we take full advantage of partial annotation learning to model the distribution of entity annotations based on all the possible label paths, avoiding being manipulated by wrong signals of false negatives. And before self-training, we use the Coca strategy to automatically calculate class-wise confidence thresholds, taking into consideration different confidence scales among label types. In the second step, to reduce the number of possible label paths in the partial CRF, we integrate the PubMedBERT-Partial-CRF model into the  Teacher-Student self-training network to iteratively estimate confidence scores and re-annotate data at each iteration.

Specifically, for each given token sequence $\boldsymbol{X}^{(k)}$, the partial CRF model will output a confidence score $s^{(k)}$ for each predicted label sequence $\boldsymbol{\hat{Y}}^{(k)}$ and choose the label sequence with the highest confidence score as the final label sequence. To get the confidence score at each token position, we use the dynamic programming forward-backward algorithm, also known as Baum–Welch algorithm \citep{baum1972inequality} to calculate the marginal distribution and derive the token-level confidence score. Formally, for each given token sequence $\boldsymbol{X}^{(k)}$, we use PubMedBERT to encode the sequence and get the hidden state of the last layer at each token position as the contextual language representation:

\begin{equation}
    h_{1:n}=\mathrm{BERT}\left(\boldsymbol{X}^{(k)};\theta_{\mathrm{BERT}}\right)
\end{equation}

Atop the contextual representation for each token, we use a linear layer to get the independent confidence score at each token position for each label, which is known as the emission score, noting that this score doesn't consider the global sequence consecutively.
\begin{equation}
\phi\left(i,y_i\right)=\mathrm{Linear}\left(h_i\right)
\end{equation}

We summarize the emission score and the transition score to get the log potentials, which can be then used to model the conditional probability of the CRF model.
\begin{equation}
\phi\left(i,y_i,y_{i+1}\right)=\phi\left(i,y_i\right)+T_{y_i,y_{i+1}}
\end{equation}

\begin{equation}
\begin{aligned}
p(\boldsymbol{Y}^{(k)}|\boldsymbol{X}^{(k)};\theta)&=\frac{\phi(\boldsymbol{Y}^{(k)})}{Z} \\&=\frac{\exp\{\sum_{i=1}^{n-1}\phi(i,y_i,y_{i+1})+\phi(n,y_{n})\}}{Z}
\end{aligned}
\end{equation}

\begin{equation}
    \alpha,Z=\mathrm{Forward}\left(\phi\right)
\end{equation}
\begin{equation}
    \beta,Z=\mathrm{Backward}\left(\phi\right)
\end{equation}
where $Z$ denotes the partition function, $\mathrm{Forward}(\cdot)$ and $\mathrm{Backward}(\cdot)$ denote the forward and backward algorithms separately, $\alpha$ and $\beta$ denote the variables of the forward and backward algorithms respectively.
Finally, we can get the confidence score for the label prediction at each token position by calculating the marginal probabilities.

\begin{equation}
s_i=p(\hat{y_i}|\boldsymbol{X}^{(k)})=\frac{\alpha_i\beta_i}{Z}
\end{equation}
where $s_i$ denote the confidence score for the model to predict $\hat{y_i}$ at the $i$-th position, $\alpha_{i}$ and $\beta_{i}$ denotes the forward and backward probability at the $i$-th position correspondingly, noting that $s_i$ considers global dependency between consecutive labels.


Commonly, neural network-based models are trained in a supervised learning paradigm using a batch-wise learning to compute the gradients of the loss function with respect to each parameter of the model. In the training process, we maximize the marginal likelihood of the observed tags \citep{tsuboi2008training}, and approximate the parameters with Monte-Carlo estimates from mini-batches \citep{robbins1951stochastic}, see Equation (10). In each batch, the algorithm computes the gradients for a subset of the training data, rather than the entire dataset, in order to reduce the computational cost of the optimization. By iteratively updating the parameters in this way, the algorithm aims to find the optimal set of parameters that minimize the loss function and enable the model to make accurate predictions.

\begin{equation}
L_{\mathrm{PAL}}\left(\theta;\boldsymbol{D_p}\right)=-\sum_{k}{\log\sum_{y\in \boldsymbol{C}\left({\boldsymbol{Y_{p}}}^{\left(k\right)}\right)}{p\left(y\middle|\mathbf{X}^{\left(k\right)};\theta\right)}}
\end{equation}

In addition, the expected entity ratio is shown to be a piece of crucial prior knowledge for guiding the model to simulate the real distribution from incomplete annotations. \cite{effland2021partially} assumed that the number of named entity tags (versus O tags) over the entire distribution of sentences occur at relatively stable rates for different named entity datasets with the same task specification. So they proposed using the Expected Entity Ratio (EER) loss in conjunction with the Partial Annotation Learning (PAL) loss to conduct multi-task learning. Specifically, if there are $N$ entity annotations in total for the partially annotated dataset, the entity ratio under complete annotation is expected to be $\hat{\rho}_\theta$.  During neural network training, their practice is to sample $B$ instances from the $N$ population and to encourage the tag marginals of being part of an entity span $p(y\neq \mathrm{O})$ in each sampled batch  under the model to match the given EER, up to a margin of uncertainty $\gamma$. 

While the entity distribution in NER datasets is highly skewed, with a large proportion of ``O'' labels denoting non-entities, the overall entity distribution across the entire dataset may differ from that of individual batches. We speculate that solely relying on the entity ratio within a batch may lead to high variance and thus, limited generalization performance. As a result, we propose the Overall Expected Entity Ratio (OEER) loss to tackle this problem by exploiting an additional constraint to the EER loss to encourage the $p(y\neq \mathrm{O})$ in each sampled batch to match the average tag marginals of being entity tags under the model during the previous training batches. Specifically, during neural network training, we naturally model entity annotations in each sampled batch and also the previous batches following an entity distribution parameterized by the expected entity ratio. 
\begin{equation}
\begin{split}
L_\mathrm{OEER}(\theta;\boldsymbol{D_p},\rho_B,\rho_O,\gamma,\lambda_{B},\lambda_{O})=\lambda_{B}\max\{0,|\rho_B-\hat{\rho}_\theta|-\gamma\}\\+\lambda_{O}|\rho_B-\rho_O|
\end{split}
\end{equation}
where $\rho_B$ denotes the entity ratio under the model in the sampled batch and $\rho_O$ denotes the averaged overall entity ratio under the model during the previous training batches, $\lambda_{B}$ and $\lambda_{O}$ are balancing coefficients.

The introduction of OEER loss is primarily aimed at enhancing  the model’s recall of missing entities. To control the precision of the model,  we propose the Self-Training (ST) loss, which incorporates the expected entity ratio into our self-training framework. The motivation behind this is to balance the recall and precision in the sampled batch by aligning the entity ratio of reliable labels under the model with the entity ratio of all labels under the model. By leveraging the expected entity ratio, the entity ratio under the model is encouraged to closely approximate it, thereby ensuring that the reliable entity ratio under the model is also consistent with the expected entity ratio. This allows us to control the quality of the updated labels generated by self-training. 
\begin{equation}
L_\mathrm{ST}(\theta;\boldsymbol{D_p},\rho_B,\rho_H,\lambda_{S})=\lambda_{S}|\rho_B-\rho_H|
\end{equation}
where $\rho_H$ denotes the high-confidence entity ratio under the model and $\lambda_{S}$ is the balancing coefficient of this loss.
\begin{equation}
L(\theta;\boldsymbol{D_p},\rho_B,\rho_O,\rho_H,\gamma,\lambda_{B},\lambda_{O},\lambda_{S})=L_\mathrm{PAL}+L_\mathrm{OEER}+L_\mathrm{ST}
\end{equation}
The final loss, presented in Equation (13), combines Equation (10), Equation (11), and Equation (12), noting that the $L_\mathrm{ST}$ is zero before self-training. 




%
\begin{table}
  \caption{Statistics of the corpora}
  \centering
\resizebox{\linewidth}{!}{%
\begin{tabular}{|r|r|r|r|r|r|r|}
\hline
Entity Type                 & Split & Sentence & Annotation & Surface Form & Token    & Entity Tag \\ \hline
\multirow{3}{*}{Genes}      & train  & 74,905    & 82,803      & 28,109        & 2,402,548 & 162,012     \\ \cline{2-7} 
                            & dev    & 12,620    & 14,442      & 6,531         & 413,643   & 28,129      \\ \cline{2-7} 
                            & test   & 36,662    & 42,386      & 16,319        & 1,184,448 & 83,225      \\ \hline
\multirow{3}{*}{Chemicals}  & train  & 99,037    & 114,575     & 29,908        & 2,873,844 & 211,044     \\ \cline{2-7} 
                            & dev    & 16,201    & 18,392      & 6,863         & 470,425   & 33,890      \\ \cline{2-7} 
                            & test   & 49,766    & 56,596      & 16,834        & 1,445,364 & 102,633     \\ \hline
\multirow{3}{*}{Diseases}   & train  & 16,895    & 14,216      & 4,238         & 417,305   & 25,780      \\ \cline{2-7} 
                            & dev    & 2,754     & 2,337       & 1,009         & 69,032    & 4,174       \\ \cline{2-7} 
                            & test   & 8,738     & 7,494       & 2,564         & 217,589   & 13,320      \\ \hline
\multirow{3}{*}{Cell Lines} & train  & 12,592    & 2,500       & 1,419         & 364,398   & 8,522       \\ \cline{2-7} 
                            & dev    & 2,001     & 450         & 290           & 60,659    & 1,406       \\ \cline{2-7} 
                            & test   & 6,146     & 1,248       & 797           & 177,994   & 4,308       \\ \hline
\multirow{3}{*}{Species}    & train  & 15,195    & 5,290       & 1,567         & 451,349   & 8,286       \\ \cline{2-7} 
                            & dev    & 2,555     & 811         & 286           & 75,284    & 1,253       \\ \cline{2-7} 
                            & test   & 7,431     & 2,891       & 917           & 221,867   & 4,474       \\ \hline
\end{tabular}}
\end{table}
\vspace{-0.4cm}
\subsection{Competitor systems} 

In this study,  we compare the TS-PubMedBERT-Partial-CRF model to two types of competitors to explore the effectiveness of our proposed model for the BioNER task: fully annotated learning-based systems, and partial annotation learning-based NER systems.

For the full annotation learning-based system, we experiment with PubMedBERT (Abstract+Fulltext) \citep{gu2021domain}, which is the state-of-the-art pre-trained language model in the biomedical domain.  PubMedBERT was pre-trained from scratch using the abstracts and full-text articles from PubMed Central, which has been shown to outperform BERT \citep{devlin2018bert}, RoBERTa \citep{liu2019roberta}, BioBERT \citep{lee2019biobert}, SciBERT \citep{beltagy2019scibert} etc. on NER task based on medical language. The PubMedBERT NER tagger uses the final hidden representation of PubMedBERT for each token and inputs it into a classification layer to categorize the tokens into different NER labels.

For partial annotation learning-based systems, we implement two models from prior work. One of them is the BiLSTM-Partial-CRF model, proposed by \cite{jie2019better}, which is a commonly used baseline model for partial annotation learning. This model is based on a BiLSTM-Partial-CRF followed by a self-training framework with cross-validation. This architecture combines a recurrent neural network with a Long Short-Term Memory (LSTM) \citep{hochreiter1997long}, for learning correlations of features over the input text and a partial CRF for predicting the tag sequence. The loss function during training will be marginalized over the labels in the missing position to consider all potential labels. The trained model is used to update gold labeling distributions in the final fold by processing all but one fold and this process will continue until convergence. Another model is the EER-BERT model, proposed by \cite{effland2021partially}, which is the state-of-the-art partial annotation learning model, exceeding the results from \cite{li2021empirical} on 7 datasets. They integrated the Expected Entity Ratio loss, based on the assumption that the number of named entity tags over the entire distribution of sentences occur at relatively stable proportion. 

We note that we chose the state-of-the-art biomedical language model PubMedBERT as the backbone model for all the pre-trained language model-based competitor systems. For example, we adapted the EER-BERT model to the EER-PubMedBERT model to convert the original RoBERTa to PubMedBERT, to make the comparison as fair as possible.

\vspace{-0.4cm}
\subsection{Corpora compilation and pre-processing}
Training deep architectures usually requires large amounts of annotated gold standard data, posing a problem to applications in the biomedical domain, where corpora sometimes contain less than 500 sentences. And they are  varying in dataset size, entity distribution, genre (e.g. patents vs. scientific articles), and text type (e.g. abstract vs. full text). In order to obtain solid evaluation results, we compile the small gold standard data sets into a large collection of biomedical NER datasets following HunFlair \citep{weber2021hunflair}.

Specifically, we take into consideration the influence of corpus size and other implicit factors such as the distribution of entity mentions, and sentence length, and performed our evaluations on five entity types: genes/proteins, chemicals, diseases, cell lines, and species. Weber et al.  successively proposed HUNER \citep{weber2020huner} and HunFlair \citep{weber2021hunflair}, stand-alone biomedical entity recognition taggers covering the above-mentioned entity types. Following their work, we integrate 15 gold-standard biomedical NER corpora using a consistent format to construct the fully annotated datasets for our experiments, excluding eight corpora \footnote{Arizona Disease, BioInfer, CLL, GELLUS, IEPA, LINNEAUS, Osiris v1.2, Variome} that we don’t have access to, and the BioSemantics corpus which contains a large number of very long sentences.

Analogously to the data preprocessing pipeline of HUNER for each entity type, we aggregate the corresponding corpora which contain annotations for the respective entity type to learn a type-specific model and convert them into the standard CoNLL2003 format. In addition, we re-use the train/dev/test splits introduced by HUNER to split each resulting corpora for each entity type with a ratio of 60:10:30 among the splits. Subsequently, we convert the BIO encoding scheme in the standard CoNLL2003 format into BILOU (beginning, inside, last, outside, unit) encoding scheme, which was observed to outperform the widely adopted BIO encoding scheme for NER.  We note that splitting is carried out in a deterministic way and there is no overlap among them across corpora for the same entity type to avoid knowledge leaks.

Table 1 highlights important statistics of the corpora for the five entity types. As we can see, the distribution of corpora varies among different entity types. Take training corpora as an example. The corpora size varies between 12,592 sentences for cell lines and 99,037 sentences for chemicals. The number of entity annotations varies between 2,500 for cell lines and 114,575 for chemicals. The number of surface forms varies between 1,419 for cell lines and 29,908 for chemicals. The distribution of the number of tokens in sentences among the training corpora for five entity types can be found in supplementary materials (Supplementary Fig. S1).

\subsection{Partial annotation scenarios simulation}

In this research, one of our goals is to explore the capability of partial annotation learning models to effectively mitigate the unlabeled entity problem \citep{li2021empirical}. We assume that partial annotations for the NER task can be obtained by removing entity annotations from the fully annotated datasets and consider two entity removal schemes to simulate unlabeled entity problems in real-world situations. Noted that we only remove entity annotations in the train set and keep the golden truth dev set and test set to acquire accurate evaluation results. 

The first scheme is ``Remove Annotations Randomly'' (RAR), previously used by \cite{jie2019better,li2021empirical}, which drops entity annotations uniformly at random. By setting the entity removal rate $r$, we control the number of removal entity annotations. For example, $r=0.1$ means that we remove 10\% of all entity annotations randomly in the dataset and keep 90\% of entity annotations. The drawback of this scheme is that the entity removal process is incomplete, with a diverse set of surface forms of the removal entity annotations still occurring in the dataset, which is not realistic under certain circumstances \citep{effland2021partially}.

The second scheme is ``Remove All Annotations for Randomly Selected Surface Forms'' (RSFR), which is used by \cite{mayhew2019named} and is a more realistic yet more challenging scheme to learn from. RSFR scheme can be regarded as a simulation of distantly supervised NER, wherein entity mentions not occurring in the dictionary will consistently not be annotated in the whole dataset. To simulate data for this scenario, we group annotations by their surface forms and randomly select groups of annotations to remove, as the literal meaning of this scheme. To allow for a fair comparison with the RAR scheme, we downsample annotations grouped by surface forms until the number of removed entity annotations is roughly the same under both schemes at the same entity removal rate. Fig. 3 provides a schematic comparison of these two schemes as an illustration. The pseudo-codes for the RAR and RSFR algorithm, and the changing of the number of annotations in the dataset along entity removal rate can be found in the supplementary materials (Section B).

\begin{figure*}
  \centering
  \includegraphics[width=\textwidth]{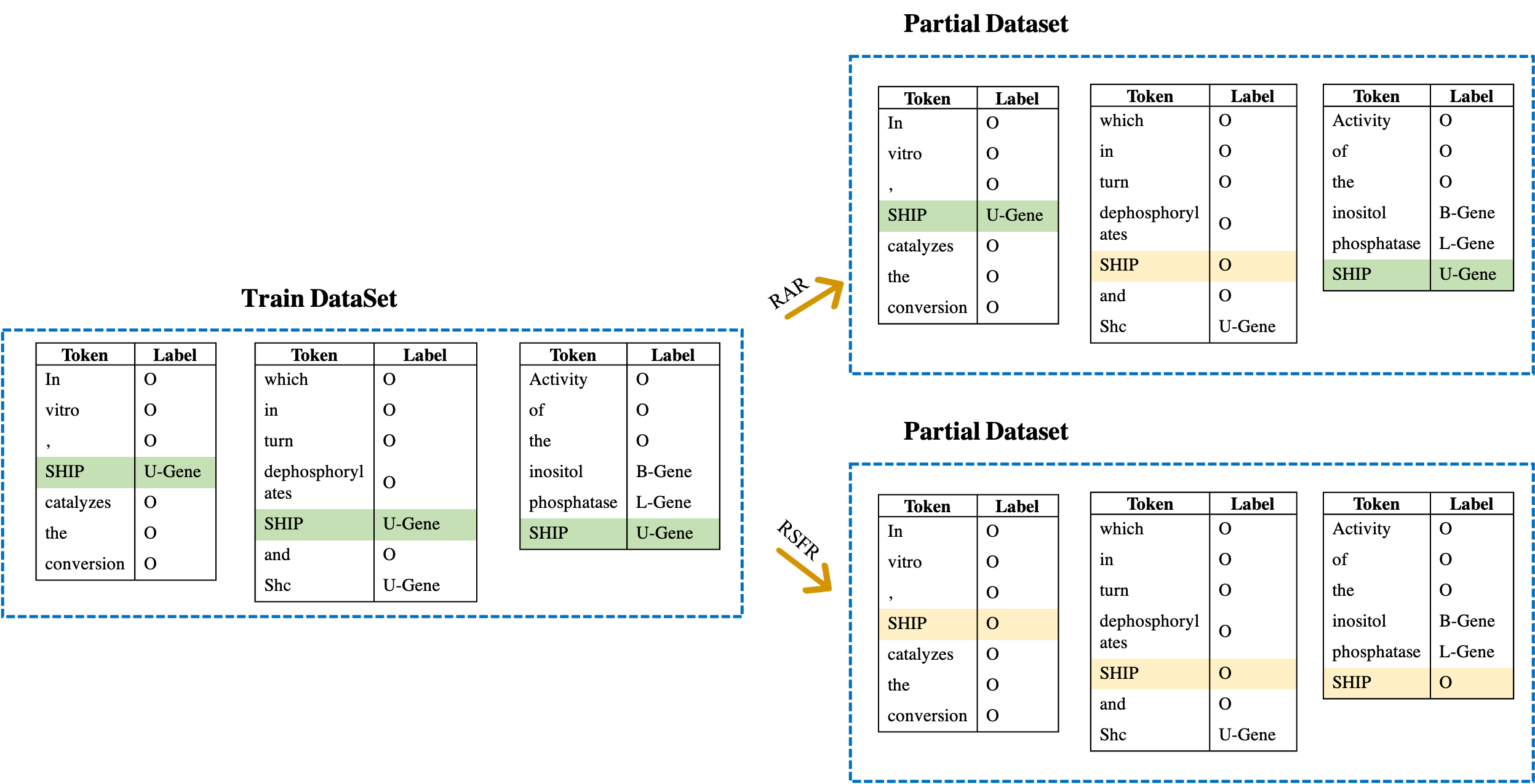}
  \caption{Comparison of two entity removal schemes (RAR and RSFR).}
\end{figure*}

\begin{figure*}
  \centering
  \includegraphics[width=\textwidth]{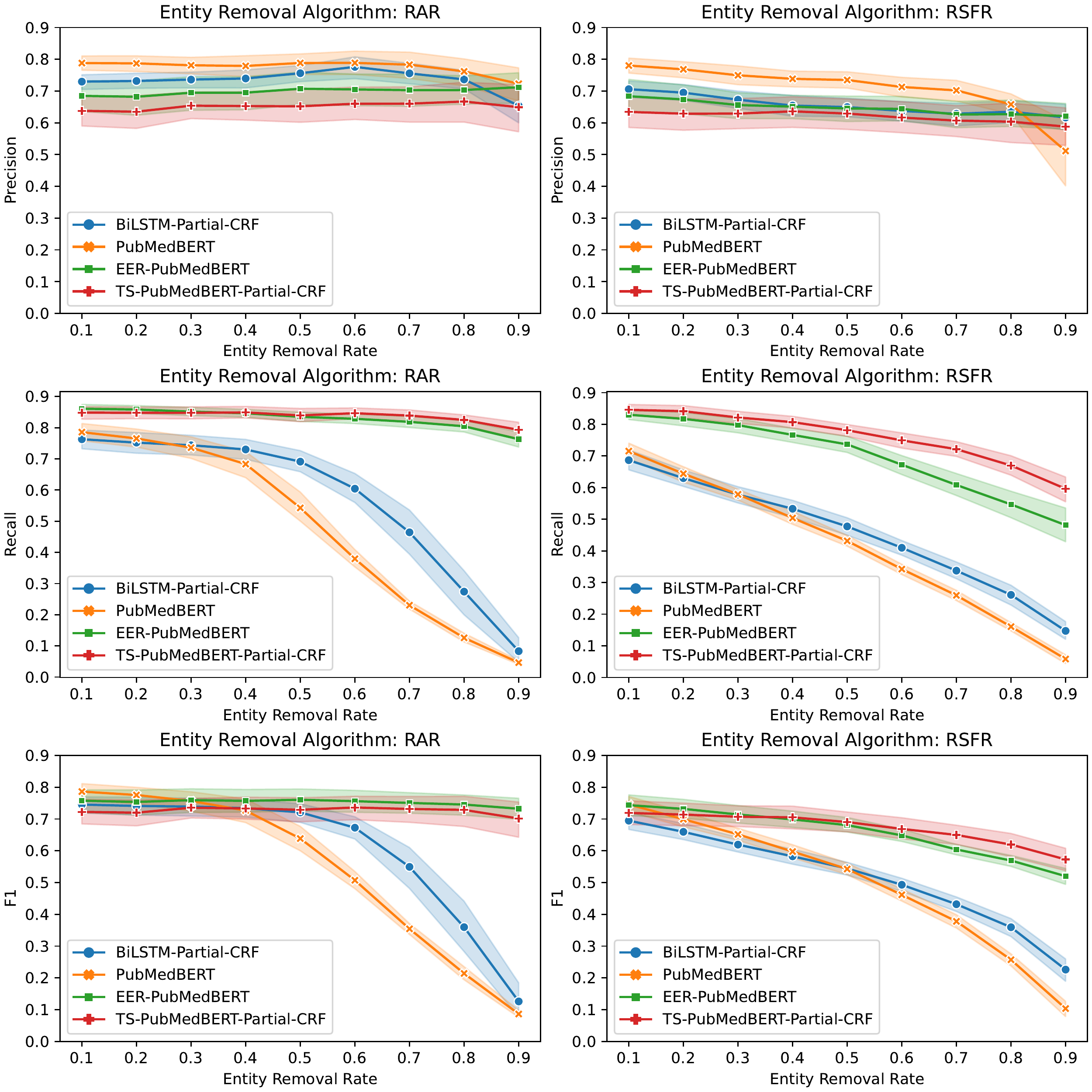}
  \caption{The overall Precision, Recall, and F1-score on the test set aggregated over all entity types. The whole figure contains 6 (3 rows × 2 columns) subplots, where each column is the group of the results for the corresponding entity removal scheme. In each subplot, the horizontal axis denotes the entity removal rate, and the vertical axis denotes the underlying evaluation metric.}
\end{figure*}

\vspace{-0.8cm}
\section{Experiments}

To verify the adaptiveness of our proposed method, we conduct  experiments based on the combination of three settings: entity type, entity removal rate, and entity removal scheme, detailed below.

\vspace{-0.4cm}
\subsection{Experimental design}

For each entity type, we generated synthetic datasets from the original fully annotated dataset by randomly removing entity annotations based on the combination of entity removal rate and entity removal scheme. The entity removal rate $r$ was set as 0.1, 0.2, …, 0.9. For each combination, we removed entity annotations from the original dataset with five different random seeds to account for the variance in model performance over different runs. In this way, 9 (entity removal rate) × 2 (entity removal scheme) × 5 (random seed) × 5 (entity type) × 4 (model) = 1,800 experiments were conducted. Furthermore, for each entity type, we applied a PubMedBERT tagger on the original fully annotated dataset to provide an upper-bound model performance, which we did not expect any of the other methods to outperform. In our work, we use the entity-level precision, recall and F1-score as the evaluation metrics to evaluate model performance, which means that the model performance is measured by its ability to correctly identify the boundary of entities and classify them in their category. The detailed experimental settings can be found in supplementary materials (Supplementary section B).

\vspace{-0.4cm}
\subsection{Overall Results}

Fig. 4 displays the comprehensive results of our empirical investigations, wherein the results are aggregated across five entity types and averaged over five independent random seeds. We have utilized a 95\% confidence interval to ensure the robustness of our findings. In order to classify the results obtained, a removal rate of 0.5 is employed, resulting in two distinct groups: a low removal rate group (0.1-0.5) and a high removal rate group (0.6-0.9). Model performance is then evaluated for both of these groups under two entity removal schemes, as shown in Table 2. Additionally, to provide detailed insights into the model's performance on each entity type, we have also recorded the corresponding F1-score, precision, and recall in (Supplementary Fig. S3, Fig. S4, Fig. S5).  We provide the full experimental results in (Supplementary Table S1.).Based on these results, we can make the following observations. 

Firstly, the main finding from our empirical experiments is that the partial annotation learning model can mitigate the unlabeled entity problem effectively. Even as the number of missing entity annotations increases, the performance degradation of our model is not extreme under either the RAR or the RSFR schemes. As we can see from Fig. 4, the F1-score of partial annotation learning models outperforms that of the full annotation learning model on average by a large margin, especially in those settings with larger entity missing rates. There is a negative correlation between the F1 score and the entity removal rate for the PubMedBERT tagger. The F1 score of the PubMedBERT tagger decreases rapidly as the entity removal rate increases to a certain extent, suggesting that the full annotation model is not robust to partial annotation even with the powerful pre-trained language model, which contains prior knowledge from millions of unlabeled data. For instance, on the Diseases dataset and under the RAR scheme, the F1 score for PubMedBERT tagger declines nearly linearly to lower than 10\% when the entity removal rate is above 0.4, while the degradation of our model from entity removal rate 0.1 to 0.9 is only 3.43\%.

Secondly, our approach is effective to eliminate the misguidance brought by unlabeled entities, surpassing the prior state-of-the-art full annotation learning PubMedBERT tagger under both entity removal schemes. As we can see from Table 2, our model significantly outperforms the PubMedBERT tagger, especially in high removal rates. Under the RAR scheme, our model achieves 72.46\% F1 score for high removal rate group, outperforming PubMedBERT tagger by 43.41\%. Taking the results for one entity type as an example, on the Chemicals dataset, and under the RAR scheme, our F1 score exceeds that of PubMedBERT tagger by 8.34\% when the removal rate is 0.5, 48.51\% when the removal rate is 0.7 and 75.23\% when the removal rate is 0.9. On the Species dataset, our model can achieve the F1 score of 65.53\% under the RAR scheme and 56.88\% under the RSFR scheme with only nearly 800 entity annotations (entity removal rate is equal to 0.9), whereas the PubMedBERT tagger can't acquire enough information to train the model, getting F1 score of 9.46\% and 15.09\% correspondingly. 

Thirdly, compared with the commonly used partial annotation learning model BiLSTM-Partial-CRF, our model exhibits a widening gap in the F1 score as the entity removal ratio increases. Taking the Disease dataset as an example, under the RAR scheme, the F1 score gap between our model and the BiLSTM-Partial-CRF model is 8.48\% when the entity removal rate is 0.7 and 64.82\%  when the entity removal rate is 0.9. This suggests that adopting a pretraining language model as the encoder for a partial annotation learning model might play an essential role in mitigating the unlabeled entity problem under high entity missing rates. Further,  as we can see from Supplementary Fig. S3, our model performs on par with the state-of-the-art partial annotation learning model EER-PubMedBERT on average, while achieving overall better results under the RSFR scheme except on Cell Lines dataset. There are modest improvements compared to the EER-PubMedBERT model under the RSFR scheme, especially on a high missing entity ratio. For example, on the disease dataset, our model outperforms EER-PubMedBERT by 16.52\% when the entity removal rate is equal to 0.9.


Fourthly, the RSFR entity removal scheme, which is more realistic, is also more challenging compared to the RAR scheme. For the RAR scheme, the partial annotation learning model can mitigate the unlabeled entity problem well when the entity removal rate is small, while the model performance starts to decline at the initial stage of removing entity annotations with the RSFR scheme. Fig. 4 demonstrates a stronger relationship between the removal rate and F1 score for the RSFR scheme, which has a steeper slope compared with the RAR scheme, indicating that even a small ratio of missing entity annotations can increase the difficulty of model learning. On the Diseases dataset and in entity removal rate 0.9, our model can acquire the F1 score of 80.32\% under the RAR scheme which is almost close to the upper bond 85.17\%, while only 71.71\% for the RSFR scheme. In addition, \cite{mayhew2019named} proposed adding false positive labels as noise to explore the effectiveness of the partial annotation learning model under this scenario, which is also valuable to explore. We will leave this investigation to future work.

Finally, as we can see from Fig.4, the F1 score curve of each subplot is roughly similar in shape to that of recall, and the precision remains relatively high even for full annotation learning model under high removal rate, suggesting that the model performance is mostly decided by the recall. The advantage of partial annotation learning models is that even with a large number of missing entity annotations, they can circumvent all possible unlabeled entities to achieve high recall with a little sacrifice to precision compared to the full annotation learning model.



\begin{table}
\centering
\caption{Results grouped by entity removal rate}
\resizebox{\linewidth}{!}{%
\begin{tabular}{|l|l|l|l|l|l|}
\hline
  Rate & Scheme & Model &            P &           R &          F1 \\
\hline
\multirow{8}{*}{0.1-0.5} & \multirow{4}{*}{RAR} & BiLSTM-Partial-CRF &  73.85±1.15 &  73.60±1.51 &  \textbf{73.62±1.26} \\
        &      & EER-PubMedBERT &  69.27±2.36 &  \textbf{85.04±0.63} &  \textbf{75.77±1.65} \\
        &      & PubMedBERT &  \textbf{78.47±1.32} &  70.26±2.25 &  \textbf{73.68±1.72} \\
        &      & Our Model &  64.62±2.17 &  \textbf{84.64±0.90} &  \textbf{72.81±1.70} \\
\cline{2-6}
        & \multirow{4}{*}{RSFR} & BiLSTM-Partial-CRF &  67.54±1.34 &  58.09±1.79 &  62.02±1.43 \\
        &      & EER-PubMedBERT &  66.18±1.96 &  78.96±1.15 &  \textbf{71.39±1.33} \\
        &      & PubMedBERT &  \textbf{75.42±1.21} &  57.46±2.01 &  64.75±1.59 \\
        &      & Our Model &  63.14±2.18 &  \textbf{81.93±0.94} &  \textbf{70.72±1.60} \\
\hline
\multirow{8}{*}{0.6-0.9} & \multirow{4}{*}{RAR} & BiLSTM-Partial-CRF &  \textbf{73.05±2.13} &  35.64±4.90 &  42.69±5.06 \\
        &      & EER-PubMedBERT &  70.57±2.46 &  \textbf{80.37±1.09} &  \textbf{74.62±1.68} \\
        &      & PubMedBERT &  \textbf{76.41±2.21} &  19.55±2.60 &  29.05±3.25 \\
        &      & Our Model&  65.91±3.10 &  \textbf{82.58±1.10} &  \textbf{72.46±2.35} \\
\cline{2-6}
        & \multirow{4}{*}{RSFR} & BiLSTM-Partial-CRF &  \textbf{62.94±1.81} &  28.88±2.36 &  37.76±2.40 \\
        &      & EER-PubMedBERT &  \textbf{62.96±2.05} &  57.72±2.51 &  58.56±1.36 \\
        &      & PubMedBERT &  \textbf{64.60±3.33} &  20.51±2.23 &  30.00±2.86 \\
        &      & Our Model &  \textbf{60.37±2.76} &  \textbf{68.42±1.89} &  \textbf{62.78±1.83} \\
\hline
\end{tabular}}
\end{table}

\vspace{-0.8cm}
\section{Discussion}

\subsection{Effects of partial annotation learning}

It is instructive to compare partial and full annotation learning models. For all five entity types we evaluate, partial annotation learning methods on average have better F1 scores when compared to the full annotation learning model, as the removal rate gets higher. The full annotation learning model can achieve good results when there are a few missing entity annotations, while as the number of missing entity annotations increases, the advantage of partial annotation learning models begins to emerge. As we can see from Table 2, the PubMedBERT tagger consistently achieves the best precision regardless of entity removal rate and entity removal method. While its recall drops sharply when the removal rate changes from low to high, with recall declining from 70.26\% to 19.55\% under the RAR scheme and from 57.46\% to 20.51\% under the RSFR scheme. 

The BiLSTM-Partial-CRF model, which is a partial annotation learning model without prior knowledge from pretraining and Transformer architecture, starts to show effectiveness under high removal rates. The BiLSTM-Partial-CRF model achieves a similar F1 score compared to PubMedBERT tagger under low removal rates but outperforms PubMedBERT tagger by 13.64\%  for high removal rates under the RAR scheme. As we can see from Fig. 4, the performance of BiLSTM-Partial-CRF starts to exceed that of the PubMedBERT when the entity removal rate increases to nearly 0.4. For our model and the EER-PubMedBERT model, the effectiveness of partial annotation learning starts to show even at a low removal rate, especially under the RSFR scheme. Our model improves over the PubMedBERT tagger by 5.97\% under the RSFR scheme for a low removal rate, and by 32.78\% for a high removal rate, verifying the effectiveness of partial annotation learning.

As we can see, there is still a gap between the state-of-the-art (SOTA) fully annotated learning methods and the partial annotation learning methods under low missing rates. Although significant progress has been made by partial annotation learning methods, they sacrifice precision to achieve proportionally higher recall. This indicates that it could be a promising option to construct a NER pipeline to achieve better overall model performance in practical applications, by using partial annotation learning to recall entity annotations first and then using a full annotation learning model to further improve precision.
\vspace{-0.4cm}
\subsection{Effects of entity distribution among datasets}
As we can see from Table 1, the entity distribution varies among datasets, which will affect the model's sensitivity to fluctuations in the partially annotated training data. The Cell Lines and Species datasets have relatively sparse entity distributions with a small number of training instances and fewer entity annotations. As illustrated in Supplementary Fig. S3, Fig. S4, and Fig. S5, we find that the precision scores for our model on these two datasets are severely lower than that of the upper bond, while the recall scores are notably above that of the upper bond overall. After balancing between precision and recall, F1 scores on these two datasets are relatively stable across different removal rates regardless of the entity removal schemes. While the slopes of our model F1 sore on the other three datasets under the RSFR scheme are apparently higher, suggesting that our model is more inclined to overfit on entity-dense datasets as the increase of missing entities under the RSFR scheme. This demonstrates that it's important to consider the precision/recall trade-off for partial annotation learning models. Furthermore, partial annotation learning models are sensitive to entity-sparsed datasets, which we can see from the confidence interval range. The confidence interval range is relatively higher on the Cell Lines and Species dataset compared to the other three datasets no matter the entity removal scheme used.
\vspace{-0.8cm}
\section{Conclusion}

In this work, we present the TS-PubMedBERT-Partial-CRF architecture, a NER model based on partial annotation learning designed for dealing with the unlabeled entity problem in NER, and explore its effectiveness in BioNER. Considering missing entity ratios and different annotation scenarios, we designed a set of empirical experiments on five Biomedical entities including Cell Lines, Diseases, Genes, Chemicals and Species, and compared them against three partial annotation learning models (TS-PubMedBERT-Partial-CRF, BiLSTM-Partial-CRF, EER-PubMedBERT), and the state-of-the-art full annotation learning PubMedBERT tagger. Our results strongly confirm the feasibility and usefulness of partial annotation learning, which shows a strong capacity to learn from partially annotated corpora even under extreme conditions. A limitation of our work is that we do not conduct ablation experiments to verify the effectiveness of each separate modification, e.g., self-training loss, which we intend to explore in future work. This study aims to provide insights into the best practices for partial annotation learning in biomedical information extraction and help practitioners make informed decisions when dealing with partially annotated data in real-world applications.






\vspace{-0.8cm}
\section*{Acknowledgements}
We acknowledge the support of Tian-Yuan Huang for drawing the figures.
\vspace{-0.8cm}
\section*{Funding}
This work is supported by the China Scholarship Council (CSC).

\vspace{-0.8cm}
\bibliographystyle{natbib}
\bibliography{document}

\appendix
\renewcommand{\thefigure}{S\arabic{figure}}
\renewcommand{\thetable}{S\arabic{table}}
\renewcommand{\thesection}{\Alph{section}}
\setcounter{figure}{0}
\setcounter{table}{0}
\begin{appendices}
\addcontentsline{toc}{chapter}{Appendices}

\section{Dataset Distribution}

Fig. S1 shows the distribution of the number of tokens in sentences among the training corpora for five entity types and the figure has used a base-10 logarithmic transformation on the $x$ axis.

\section{Partial Annotation Scenarios}

The changing of the number of annotations in the dataset along entity removal rate is shown in Fig. S2. We can see that RAR and RSFR have a similar number of entity annotations under the same entity removal rate. 

The pseudo-code for the RAR algorithm is shown in Algorithm 1.

\begin{algorithm}
\caption{Remove Annotations Randomly (RAR)}
\begin{algorithmic}[1]
\REQUIRE $r$ for the removal ratio; $\mathrm{Train\_Instances}$ for the instance list in the original training set, in which each instance is a collection of (TOKEN, LABEL) pairs, formatted as $\mathrm{Instance (TOKEN, LABEL)}$; $\mathrm{Entity\_Spans}$ for entity span list in the original training set, in which each entity annotation is formatted as $\mathrm{Span (NAME, LABEL, START, END, Ins\_ID)})$
\ENSURE $\mathrm{Partial\_Instances}$
\STATE $\mathrm{Num\_Removal} \gets  \mathrm{round}(r*\mathrm{length(Entity\_Spans))}$
\STATE $flag \gets 0$
\STATE Shuffle $\mathrm{Entity\_Spans}$

\FOR{each $entity \in  \mathrm{Entity\_Spans}$}
\IF {$flag$<$\mathrm{Num\_Removal}$}
    \STATE
        $\mathrm{Train\_Instances[Ins\_ID][START:END].LABEL \gets ``O"}$
    \STATE $flag \gets flag+1$
\ENDIF
\ENDFOR

\STATE $\mathrm{Partial\_Instances \gets Train\_Instances}$ 
\end{algorithmic}
\end{algorithm}

The pseudo-code for the RSFR algorithm is shown in Algorithm 2.
\begin{algorithm}
\caption{Remove All Annotations for Randomly Selected Surface Forms (RSFR)}
\begin{algorithmic}[1]
\REQUIRE $r$ for the removal ratio; $\mathrm{Train\_Instances}$ for the instance list in the original training set, in which each instance is a collection of (TOKEN, LABEL) pairs, formatted as $\mathrm{Instance (TOKEN, LABEL)}$; $\mathrm{Entity\_Spans}$ for entity span list in the original training set, in which each entity annotation is formatted as $\mathrm{Span (NAME, LABEL, START, END, Ins\_ID)})$; $\mathrm{Removal\_Name}$ for an empty set to store the removal surface forms
\ENSURE $\mathrm{Partial\_Instances}$
\STATE $\mathrm{Num\_Removal} \gets  \mathrm{round}(r*\mathrm{length(Entity\_Spans))}*0.99$
\STATE $flag \gets 0$
\STATE Shuffle $\mathrm{Entity\_Spans}$

\FOR{each $entity \in  \mathrm{Entity\_Spans}$}
    \STATE $\mathrm{Surface\_Form} \gets entity.\mathrm{NAME}$
     \STATE $\mathrm{TYPE} \gets entity.\mathrm{LABEL}$
\IF {$flag$<$\mathrm{Num\_Removal}$ \AND $\mathrm{Surface\_Form} \notin \mathrm{Removal\_Name}$}

    \STATE Add $\mathrm{Surface\_Form}$ to $\mathrm{Removal\_Name}$
    \FOR{each $span \in \mathrm{Entity\_Spans}$}
    \IF{$span.\mathrm{NAME}= \mathrm{Surface\_Form} \quad
    \AND \quad span.\mathrm{LABEL}=\mathrm{TYPE}$}

    \STATE $\mathrm{Ins\_ID} \gets span.\mathrm{Ins\_ID} $
    \STATE $\mathrm{START} \gets span.\mathrm{START} $
    \STATE $\mathrm{END} \gets span.\mathrm{END}$
    \STATE $\mathrm{LABEL} \gets span.\mathrm{LABEL}$
   \STATE
        $\mathrm{Train\_Instances[Ins\_ID][START:END].LABEL \gets  ``O"}$
    \STATE $flag \gets flag+1$
    \ENDIF
    \ENDFOR
\ENDIF
\ENDFOR
\STATE $\mathrm{Partial\_Instances \gets Train\_Instances}$ 
\end{algorithmic}
\end{algorithm}

\section{Experimental settings}

For the baseline models, we follow the officially released implementation from the authors and made a few modifications to fit our tasks. All experiments are implemented on an NVIDIA A100 GPU. For models based on PubMedBERT pre-trained language model, we use the Huggingface implementation of PubMedBERT \footnote{https://huggingface.co/microsoft/BiomedNLP-PubMedBERT-base-uncased-abstract-fulltext}. As shown in Fig. 3, majority of sentence lengths are less than 100, so we set the maximum length of the sentence as 128 for the pre-trained transformer models to ensure optimal performance.

For the BiLSTM-Partial-CRF model, we use the Pytorchimplementation\footnote{https://github.com/allanj/ner\_incomplete\_annotation} and employ 100-dimensional pre-trained word embeddings GloVe to initialize the lookup table. We train the model for 100 epochs using the stochastic gradient descent optimizer on all synthetic datasets for five entity types. The learning rate is set to 0.001, and the batch size varies among different entity types due to memory constraints. The remaining hyperparameters are left as their default value in the authors' implementations.

For the PubMedBERT-based models including the PubMedBERT tagger, the EER-PubMedBERT model, and our proposed TS-PubMedBERT-Partial-CRF model, training is done by mini-batch stochastic gradient descent (SGD) with exponential learning rate decay and the initial learning rate is set to $5e-5$. The model is trained for 50 epochs for the entity types of diseases, species, and cell lines, and 20 epochs for the entity types of chemicals and genes with early stopping. For the EER-PubMedBERT model, the default value of the hyperparameter expected entity ratio is 0.25, which can’t make the model converge. So, we use grid search to tune this hyperparameter for values in \{0.01, 0.05, 0.15, 0.25, 0.5, 0.75\}. We find that setting the expected entity ratio as 0.01 achieves the best model performance for all five entity types. For the TS-PubMedBERT-Partial-CRF model, to keep consistent with the baseline models, especially EER model for a fair comparison, we use the same hyperparameter expected entity ratio, uncertainty margin $\gamma$, and also the same balancing coefficient of EER loss $\lambda_B$. We only tune the begging step of self-training, the balancing coefficient $\lambda_O$ in overall EER loss, and the coefficient $\lambda_S$, and then report the best performance. Note that in the hyperparameter searching process, we select the hyperparameters based on the best performance on the development set of the fully annotated corpora and apply these hyperparameters to the partially annotated corpora of the corresponding entity type.

It should be also noted that selecting a single set of hyperparameters to tune the model may not be the optimal choice, as each synthetic dataset may have a distribution that differs from the fully annotated dataset. Ideally, each synthetic dataset should be tuned separately. However, given the size of our experimental setup, which consists of 1,800 experiments, it is not feasible to tune each dataset separately. To gain a comprehensive understanding of the comparison between partial annotation learning methods and full annotation learning methods, we have opted for a compromise approach, in which we apply one set of hyperparameters from the corresponding model to all combinations of entity removal rate and entity removal scheme for the same entity type. Nevertheless, model parameters can be further tuned to achieve even higher performance.

\begin{figure}
  \centering
  \includegraphics[width=\textwidth]{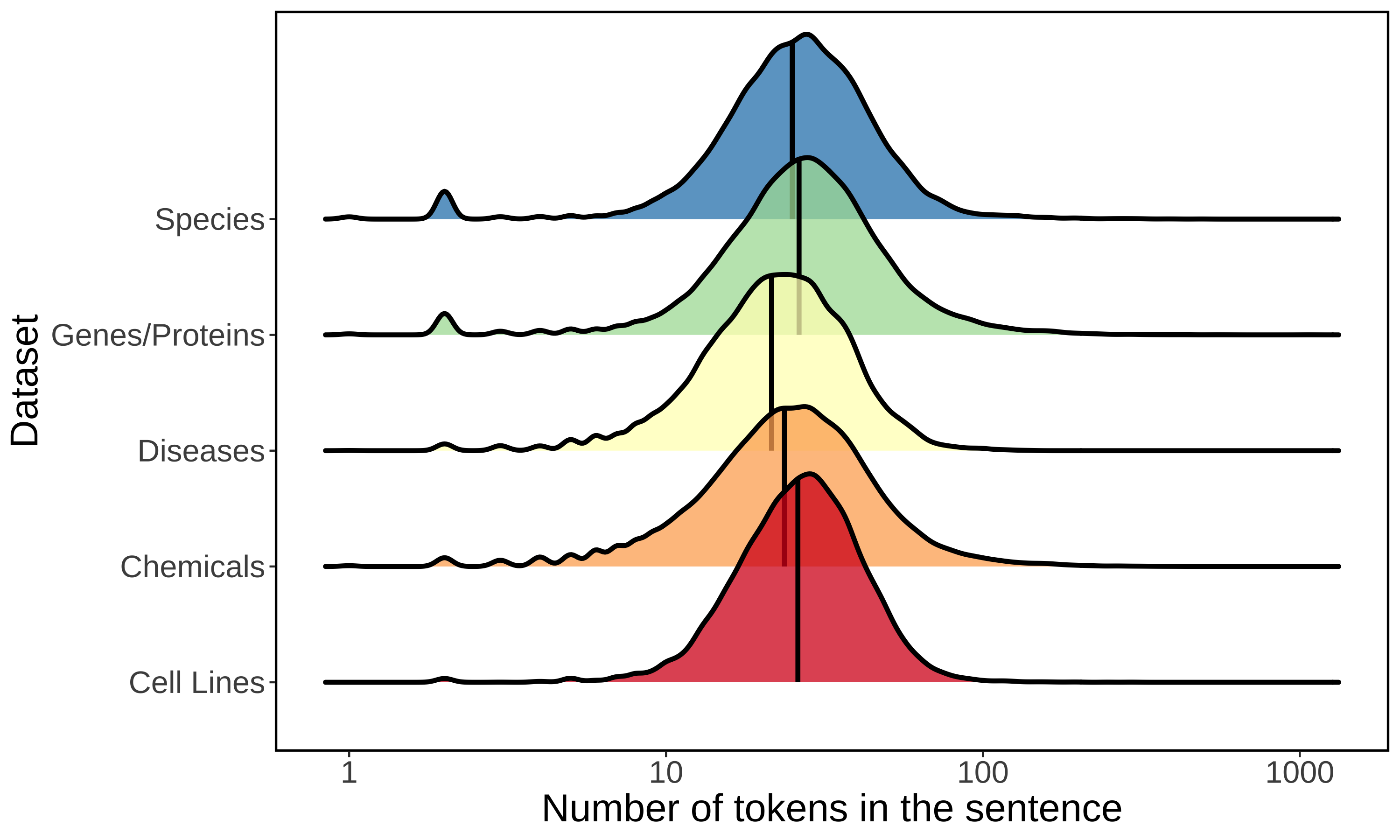}
  \caption{Distribution of the number of tokens in sentences among the training corpora.}
\end{figure}
\begin{figure*}
  \centering
  \includegraphics[width=\textwidth]{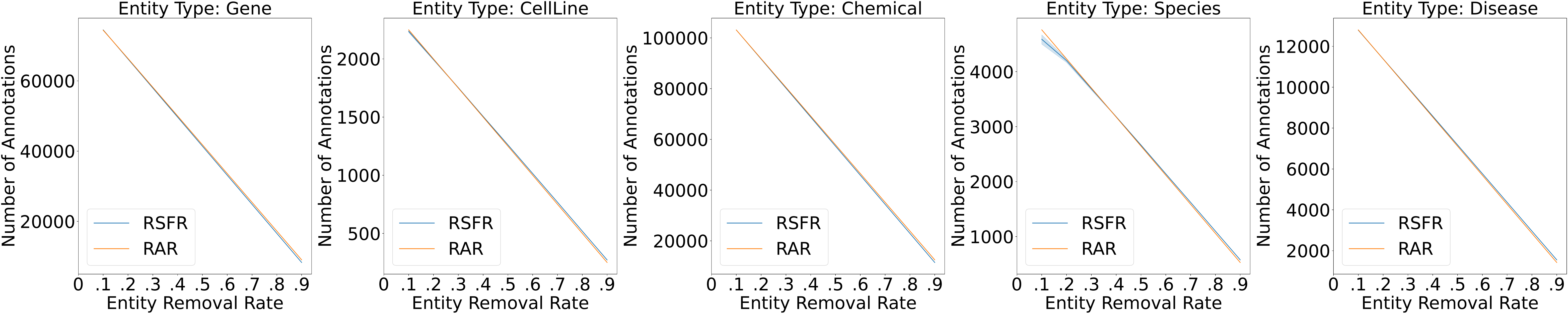}
  \caption{Distribution of the number of annotations along the entity removal rate.}
\end{figure*}

\begin{figure*}
  \centering
  \includegraphics[width=\textwidth]{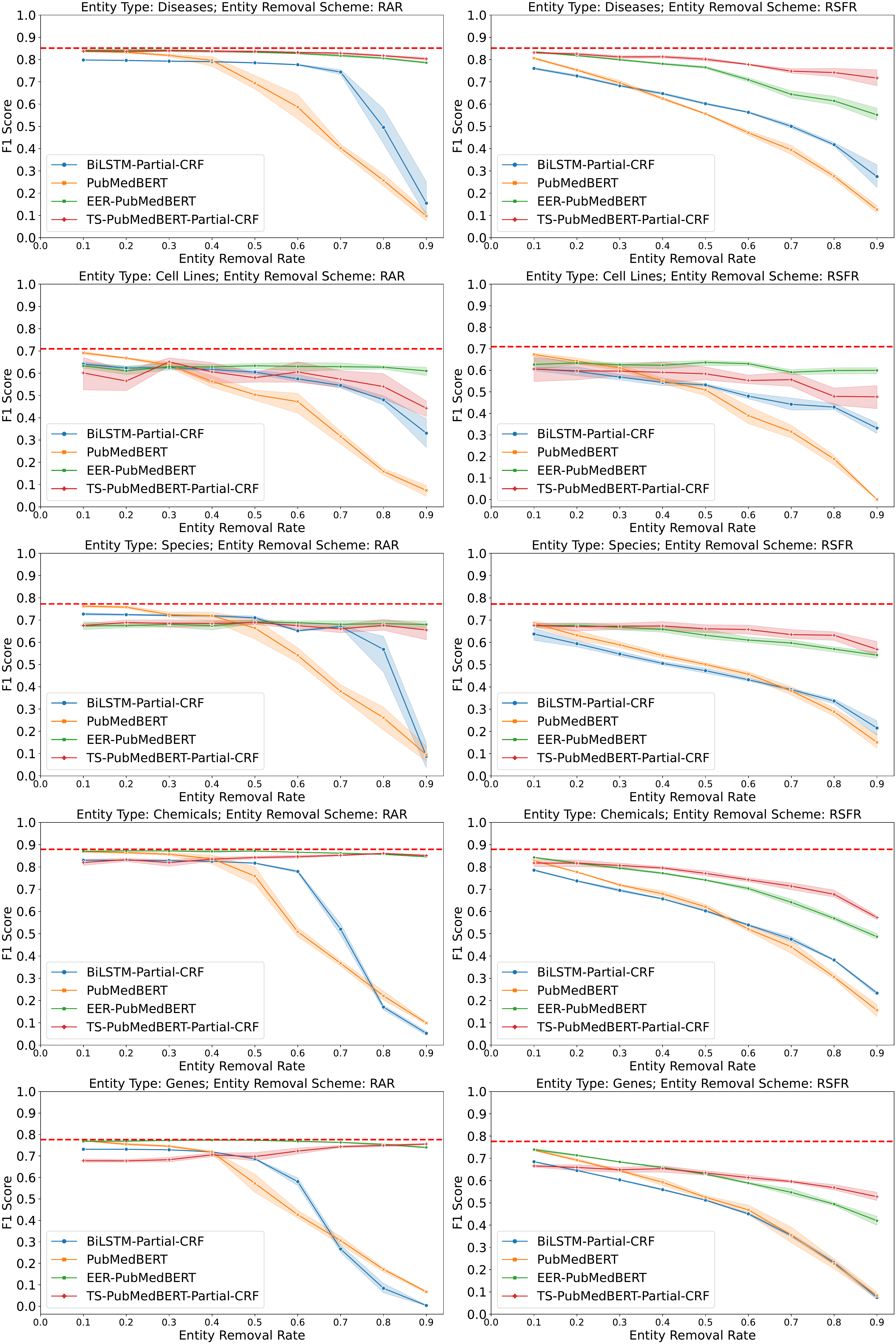}
  \caption{The F1-score on the test set. The whole figure contains 10 (5 rows × 2 columns) subplots, where each row of subplots reflects the curves of each entity type and each column of subplots reflects the curves of each entity removal scheme. In each subplot, the horizontal axis denotes the entity removal rate, the vertical axis denotes the F1-score, and the red dotted line denotes the upper bound F1-score, which is the F1-score of PubMedBERT tagger on fully annotated dataset.}
\end{figure*}

\begin{figure*}
  \centering
  \includegraphics[width=\textwidth]{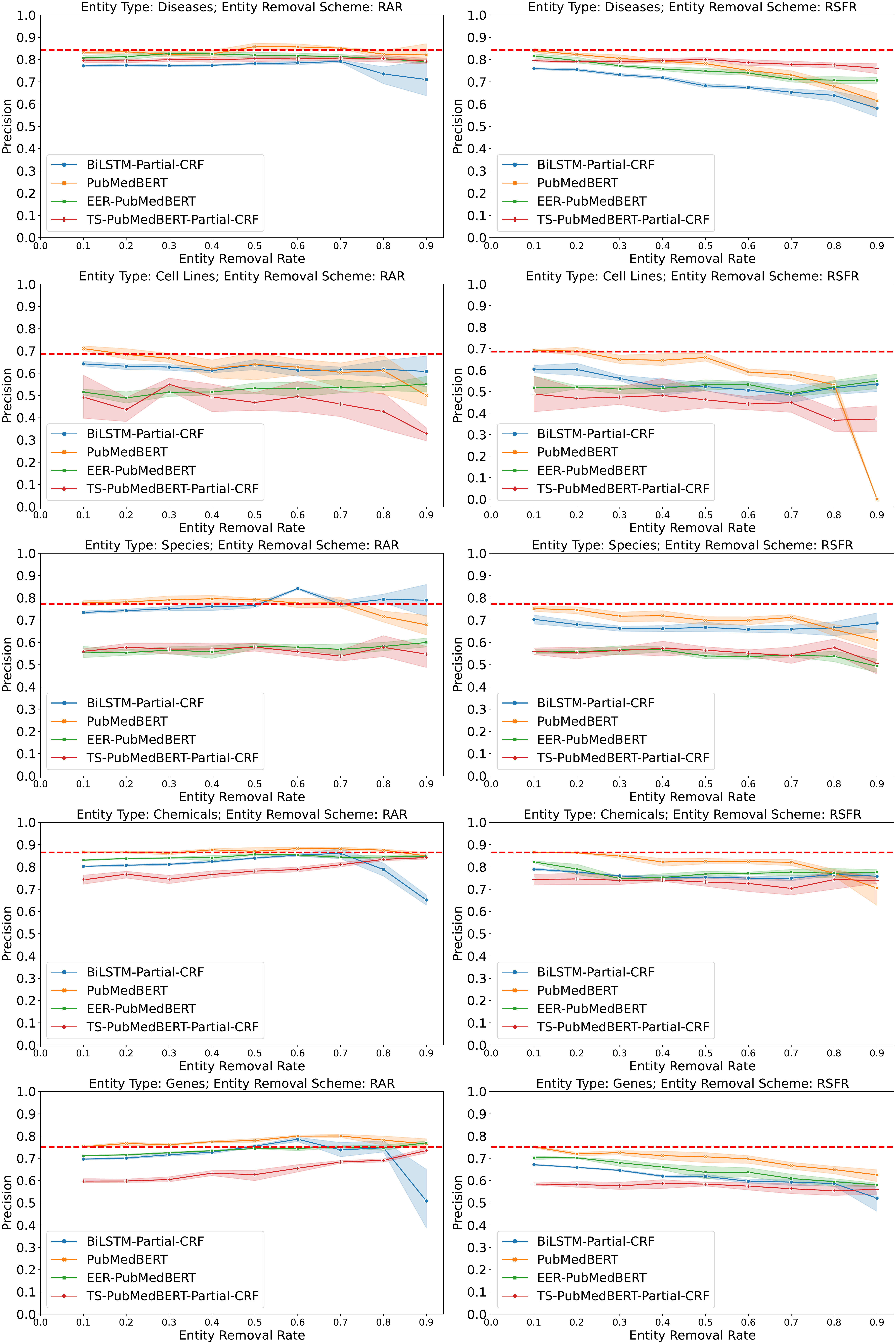}
  \caption{The Precision on the test set}
\end{figure*}

\begin{figure*}
  \centering
  \includegraphics[width=\textwidth]{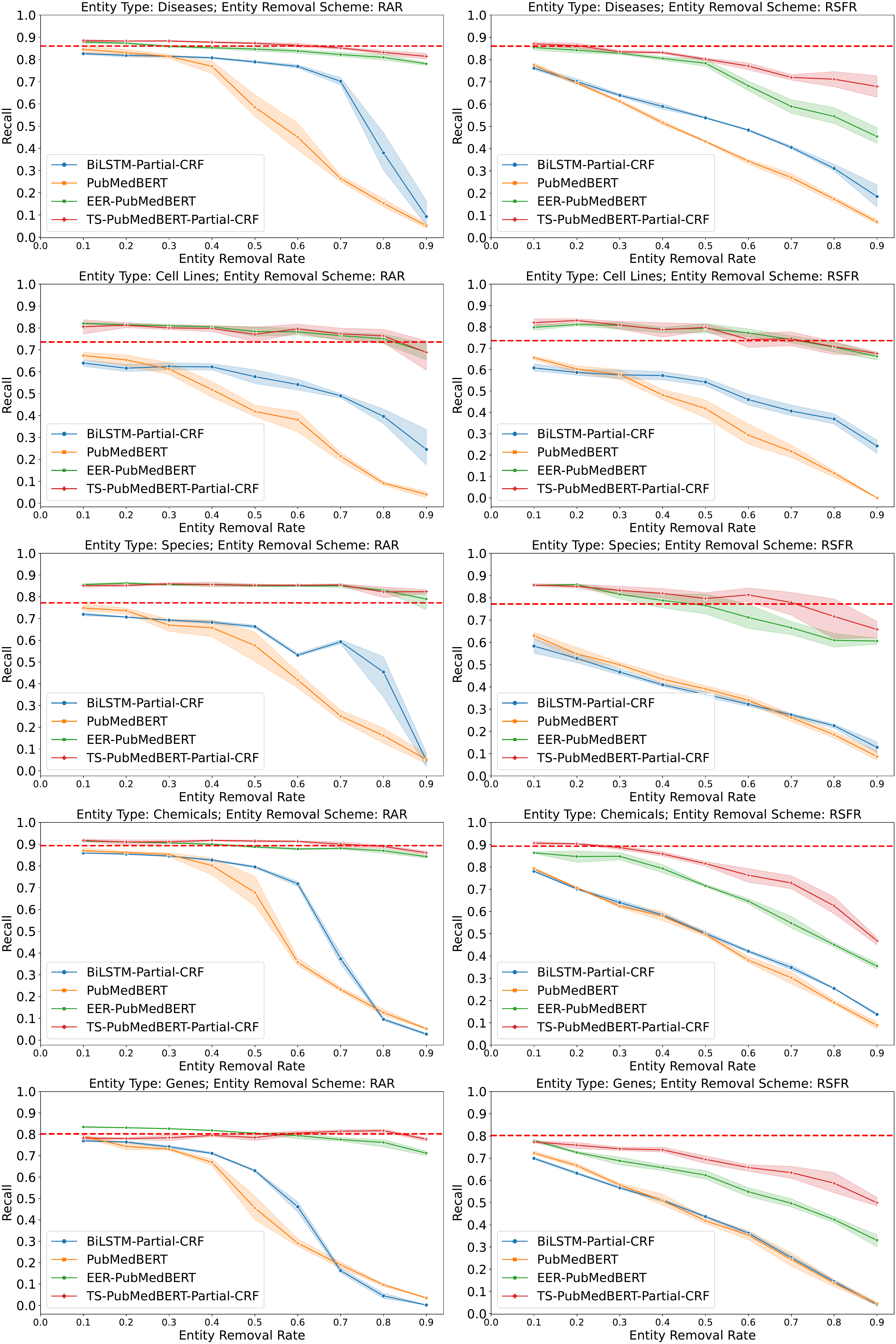}
  \caption{The Recall on the test set}
\end{figure*}

\section{Experimental Results}
To provide detailed insights into the model's performance on each entity type, we also record the corresponding F1-score, precision, and recall, as shown in Fig. S3, Fig. S4, and Fig. S5, respectively. The layouts of Fig. S4 and Fig. S5 are the same as that of Fig. S4, except for the vertical axis,  denoting precision and recall, respectively. Table. S1 presents  the full results.
\begin{center}
\tablecaption{Main Results averaged over all random seeds}
\tablefirsthead{
\hline
     Dataset & Scheme & Rate & Model&   P &            R &          F1 \\\hline
}
\tablehead{%
\multicolumn{7}{c}%
{{\bfseries  Continued from previous page}} \\
\hline
     Dataset & Scheme & Rate & Model&   P &  R &  F1 \\\hline}
%
\tabletail{ \hline}

\tablelasttail{\bottomrule}
\onecolumn
\begin{supertabular}{l|l|l|l|l|l|l}
CellLine &    RAR &   0.1 &        BiLSTM-Partial-CRF &  64.22±1.18 &  63.97±1.70 & 64.08±1.12 \\
CellLine &    RAR &   0.1 &            EER-PubMedBERT &  51.55±1.44 &  82.09±0.51 & 63.32±1.17 \\
CellLine &    RAR &   0.1 &                PubMedBERT &  71.06±1.23 &  67.39±1.85 & 69.15±0.84 \\
CellLine &    RAR &   0.1 & TS-PubMedBERT-Partial-CRF & 49.31±11.00 &  80.56±3.64 & 60.13±8.21 \\
CellLine &    RAR &   0.2 &        BiLSTM-Partial-CRF &  63.15±1.52 &  61.65±1.82 & 62.36±0.87 \\
CellLine &    RAR &   0.2 &            EER-PubMedBERT &  48.92±3.01 &  81.43±0.72 & 61.06±2.24 \\
CellLine &    RAR &   0.2 &                PubMedBERT &  68.45±2.75 &  65.45±2.83 & 66.79±0.30 \\
CellLine &    RAR &   0.2 & TS-PubMedBERT-Partial-CRF &  43.65±6.05 &  81.29±1.30 & 56.50±4.93 \\
CellLine &    RAR &   0.3 &        BiLSTM-Partial-CRF &  62.78±1.43 &  62.39±1.79 & 62.54±0.59 \\
CellLine &    RAR &   0.3 &            EER-PubMedBERT &  51.51±2.67 &  81.01±0.88 & 62.92±1.81 \\
CellLine &    RAR &   0.3 &                PubMedBERT &  66.73±2.11 &  61.24±2.94 & 63.79±1.39 \\
CellLine &    RAR &   0.3 & TS-PubMedBERT-Partial-CRF &  55.03±3.35 &  80.02±1.15 & 65.12±2.10 \\
CellLine &    RAR &   0.4 &        BiLSTM-Partial-CRF &  61.13±1.39 &  62.24±1.75 & 61.66±0.96 \\
CellLine &    RAR &   0.4 &            EER-PubMedBERT &  51.59±1.59 &  80.48±0.78 & 62.86±1.20 \\
CellLine &    RAR &   0.4 &                PubMedBERT &  62.10±4.53 &  51.84±4.23 & 56.33±3.09 \\
CellLine &    RAR &   0.4 & TS-PubMedBERT-Partial-CRF &  49.36±7.13 &  79.78±1.61 & 60.58±5.24 \\
CellLine &    RAR &   0.5 &        BiLSTM-Partial-CRF &  63.88±2.59 &  57.77±3.60 & 60.51±1.07 \\
CellLine &    RAR &   0.5 &            EER-PubMedBERT &  53.32±2.62 &  78.39±2.41 & 63.38±1.19 \\
CellLine &    RAR &   0.5 &                PubMedBERT &  63.95±6.72 &  41.88±2.81 & 50.36±2.51 \\
CellLine &    RAR &   0.5 & TS-PubMedBERT-Partial-CRF &  46.87±4.71 &  77.12±3.79 & 58.01±2.53 \\
CellLine &    RAR &   0.6 &        BiLSTM-Partial-CRF &  61.34±3.06 &  54.18±2.74 & 57.45±1.91 \\
CellLine &    RAR &   0.6 &            EER-PubMedBERT &  53.02±3.65 &  78.22±1.77 & 63.07±2.12 \\
CellLine &    RAR &   0.6 &                PubMedBERT &  62.71±4.08 &  38.07±5.34 & 47.26±5.17 \\
CellLine &    RAR &   0.6 & TS-PubMedBERT-Partial-CRF &  49.54±7.72 &  79.57±2.43 & 60.56±5.13 \\
CellLine &    RAR &   0.7 &        BiLSTM-Partial-CRF &  61.45±1.90 &  49.02±1.07 & 54.53±1.23 \\
CellLine &    RAR &   0.7 &            EER-PubMedBERT &  53.66±3.49 &  76.50±1.89 & 62.96±1.82 \\
CellLine &    RAR &   0.7 &                PubMedBERT &  60.37±6.56 &  21.46±2.42 & 31.64±3.39 \\
CellLine &    RAR &   0.7 & TS-PubMedBERT-Partial-CRF &  46.17±6.71 &  77.30±3.07 & 57.33±4.43 \\
CellLine &    RAR &   0.8 &        BiLSTM-Partial-CRF &  61.81±4.21 &  39.63±3.86 & 48.05±2.42 \\
CellLine &    RAR &   0.8 &            EER-PubMedBERT &  53.92±1.76 &  75.10±2.39 & 62.72±0.83 \\
CellLine &    RAR &   0.8 &                PubMedBERT & 61.15±10.55 &   9.16±1.03 & 15.89±1.84 \\
CellLine &    RAR &   0.8 & TS-PubMedBERT-Partial-CRF &  42.79±9.23 &  76.40±3.44 & 54.01±6.79 \\
CellLine &    RAR &   0.9 &        BiLSTM-Partial-CRF &  60.80±8.15 &  24.55±9.29 & 33.08±7.46 \\
CellLine &    RAR &   0.9 &            EER-PubMedBERT &  55.10±3.95 &  69.00±4.33 & 61.06±2.20 \\
CellLine &    RAR &   0.9 &                PubMedBERT &  50.00±5.39 &   4.02±1.57 &  7.38±2.80 \\
CellLine &    RAR &   0.9 & TS-PubMedBERT-Partial-CRF &  32.82±3.21 &  68.87±7.73 & 44.28±3.86 \\
\hline
CellLine &   RSFR &   0.1 &        BiLSTM-Partial-CRF &  60.51±2.01 &  60.82±1.88 & 60.61±0.85 \\
CellLine &   RSFR &   0.1 &            EER-PubMedBERT &  51.92±5.17 &  79.81±1.58 & 62.73±3.42 \\
CellLine &   RSFR &   0.1 &                PubMedBERT &  69.21±0.55 &  65.59±1.05 & 67.35±0.78 \\
CellLine &   RSFR &   0.1 & TS-PubMedBERT-Partial-CRF &  48.90±9.19 &  82.01±2.06 & 60.59±6.81 \\
CellLine &   RSFR &   0.2 &        BiLSTM-Partial-CRF &  60.35±3.21 &  58.69±1.53 & 59.49±2.31 \\
CellLine &   RSFR &   0.2 &            EER-PubMedBERT &  52.06±0.92 &  81.12±0.69 & 63.42±0.76 \\
CellLine &   RSFR &   0.2 &                PubMedBERT &  68.90±2.11 &  60.32±1.84 & 64.31±1.70 \\
CellLine &   RSFR &   0.2 & TS-PubMedBERT-Partial-CRF &  46.94±5.29 &  83.02±0.99 & 59.75±4.14 \\
CellLine &   RSFR &   0.3 &        BiLSTM-Partial-CRF &  56.19±1.45 &  57.56±2.49 & 56.84±1.44 \\
CellLine &   RSFR &   0.3 &            EER-PubMedBERT &  51.22±2.16 &  80.76±1.87 & 62.62±1.14 \\
CellLine &   RSFR &   0.3 &                PubMedBERT &  64.95±2.25 &  57.82±2.60 & 61.10±1.25 \\
CellLine &   RSFR &   0.3 & TS-PubMedBERT-Partial-CRF &  47.47±4.79 &  80.93±2.06 & 59.62±3.01 \\
CellLine &   RSFR &   0.4 &        BiLSTM-Partial-CRF &  52.16±4.41 &  57.24±2.21 & 54.38±1.68 \\
CellLine &   RSFR &   0.4 &            EER-PubMedBERT &  51.59±2.15 &  78.89±1.59 & 62.37±1.95 \\
CellLine &   RSFR &   0.4 &                PubMedBERT &  64.64±3.03 &  48.13±2.75 & 55.06±1.50 \\
CellLine &   RSFR &   0.4 & TS-PubMedBERT-Partial-CRF &  48.25±8.80 &  78.68±3.93 & 59.06±5.43 \\
CellLine &   RSFR &   0.5 &        BiLSTM-Partial-CRF &  52.38±2.20 &  54.21±1.93 & 53.21±0.65 \\
CellLine &   RSFR &   0.5 &            EER-PubMedBERT &  53.30±2.71 &  79.36±2.53 & 63.66±1.37 \\
CellLine &   RSFR &   0.5 &                PubMedBERT &  65.92±2.01 &  41.80±4.65 & 50.93±3.09 \\
CellLine &   RSFR &   0.5 & TS-PubMedBERT-Partial-CRF &  46.24±4.58 &  79.79±1.76 & 58.37±3.35 \\
CellLine &   RSFR &   0.6 &        BiLSTM-Partial-CRF &  50.67±4.41 &  45.98±2.70 & 47.98±1.55 \\
CellLine &   RSFR &   0.6 &            EER-PubMedBERT &  53.37±1.79 &  77.22±2.11 & 63.07±0.99 \\
CellLine &   RSFR &   0.6 &                PubMedBERT &  59.20±1.71 &  29.37±5.36 & 38.96±4.66 \\
CellLine &   RSFR &   0.6 & TS-PubMedBERT-Partial-CRF &  44.27±3.38 &  74.14±3.88 & 55.29±2.52 \\
CellLine &   RSFR &   0.7 &        BiLSTM-Partial-CRF &  48.62±4.46 &  40.64±2.81 & 44.23±3.19 \\
CellLine &   RSFR &   0.7 &            EER-PubMedBERT &  49.20±1.51 &  74.09±1.69 & 59.11±1.32 \\
CellLine &   RSFR &   0.7 &                PubMedBERT &  57.81±2.11 &  21.80±3.28 & 31.51±3.50 \\
CellLine &   RSFR &   0.7 & TS-PubMedBERT-Partial-CRF &  44.87±5.57 &  74.46±3.74 & 55.66±3.70 \\
CellLine &   RSFR &   0.8 &        BiLSTM-Partial-CRF &  51.66±3.37 &  36.91±2.68 & 42.91±1.67 \\
CellLine &   RSFR &   0.8 &            EER-PubMedBERT &  52.26±3.25 &  70.54±2.56 & 59.91±1.65 \\
CellLine &   RSFR &   0.8 &                PubMedBERT &  53.19±4.13 &  11.52±1.82 & 18.82±2.52 \\
CellLine &   RSFR &   0.8 & TS-PubMedBERT-Partial-CRF &  36.74±6.01 &  70.80±3.61 & 47.91±4.46 \\
CellLine &   RSFR &   0.9 &        BiLSTM-Partial-CRF &  53.54±3.52 &  24.26±3.24 & 33.18±3.00 \\
CellLine &   RSFR &   0.9 &            EER-PubMedBERT &  55.06±3.73 &  66.18±1.85 & 59.97±1.58 \\
CellLine &   RSFR &   0.9 &                PubMedBERT &   0.00±0.00 &   0.00±0.00 &  0.00±0.00 \\
CellLine &   RSFR &   0.9 & TS-PubMedBERT-Partial-CRF &  37.33±7.12 &  67.57±0.78 & 47.68±5.98 \\
\hline
    \hline
Chemical &    RAR &   0.1 &        BiLSTM-Partial-CRF &  80.29±0.15 &  85.96±0.07 & 83.03±0.09 \\
Chemical &    RAR &   0.1 &            EER-PubMedBERT &  83.06±0.35 &  91.43±0.11 & 87.04±0.15 \\
Chemical &    RAR &   0.1 &                PubMedBERT &  86.71±0.53 &  87.02±1.18 & 86.86±0.33 \\
Chemical &    RAR &   0.1 & TS-PubMedBERT-Partial-CRF &  74.21±2.22 &  91.75±0.71 & 82.04±1.55 \\
Chemical &    RAR &   0.2 &        BiLSTM-Partial-CRF &  80.78±0.51 &  85.47±0.31 & 83.06±0.17 \\
Chemical &    RAR &   0.2 &            EER-PubMedBERT &  83.81±0.25 &  91.01±0.12 & 87.26±0.12 \\
Chemical &    RAR &   0.2 &                PubMedBERT &  86.71±0.47 &  86.20±0.76 & 86.45±0.53 \\
Chemical &    RAR &   0.2 & TS-PubMedBERT-Partial-CRF &  76.79±1.75 &  91.02±1.41 & 83.28±0.86 \\
Chemical &    RAR &   0.3 &        BiLSTM-Partial-CRF &  81.21±0.45 &  84.57±0.18 & 82.85±0.17 \\
Chemical &    RAR &   0.3 &            EER-PubMedBERT &  84.05±0.28 &  90.61±0.22 & 87.20±0.11 \\
Chemical &    RAR &   0.3 &                PubMedBERT &  86.17±0.80 &  85.28±1.09 & 85.71±0.58 \\
Chemical &    RAR &   0.3 & TS-PubMedBERT-Partial-CRF &  74.50±2.23 &  91.11±1.03 & 81.96±1.67 \\
Chemical &    RAR &   0.4 &        BiLSTM-Partial-CRF &  82.37±0.56 &  82.72±0.40 & 82.54±0.10 \\
Chemical &    RAR &   0.4 &            EER-PubMedBERT &  84.18±1.43 &  90.00±0.46 & 86.98±0.63 \\
Chemical &    RAR &   0.4 &                PubMedBERT &  87.71±0.60 &  80.15±4.39 & 83.68±2.34 \\
Chemical &    RAR &   0.4 & TS-PubMedBERT-Partial-CRF &  76.63±1.83 &  91.77±0.33 & 83.51±1.04 \\
Chemical &    RAR &   0.5 &        BiLSTM-Partial-CRF &  84.00±0.34 &  79.56±0.53 & 81.72±0.23 \\
Chemical &    RAR &   0.5 &            EER-PubMedBERT &  85.61±0.38 &  88.77±0.50 & 87.16±0.10 \\
Chemical &    RAR &   0.5 &                PubMedBERT &  87.04±1.98 &  67.83±7.83 & 75.92±4.42 \\
Chemical &    RAR &   0.5 & TS-PubMedBERT-Partial-CRF &  78.13±1.30 &  91.45±0.58 & 84.26±0.66 \\
Chemical &    RAR &   0.6 &        BiLSTM-Partial-CRF &  85.31±0.64 &  71.83±1.41 & 77.98±0.58 \\
Chemical &    RAR &   0.6 &            EER-PubMedBERT &  85.41±0.57 &  87.82±0.59 & 86.59±0.12 \\
Chemical &    RAR &   0.6 &                PubMedBERT &  88.26±0.73 &  35.79±2.15 & 50.89±2.18 \\
Chemical &    RAR &   0.6 & TS-PubMedBERT-Partial-CRF &  78.87±1.27 &  91.31±0.35 & 84.62±0.72 \\
Chemical &    RAR &   0.7 &        BiLSTM-Partial-CRF &  86.15±2.61 &  37.35±2.89 & 52.04±2.96 \\
Chemical &    RAR &   0.7 &            EER-PubMedBERT &  84.37±0.59 &  88.08±0.29 & 86.18±0.21 \\
Chemical &    RAR &   0.7 &                PubMedBERT &  88.14±0.86 &  23.23±1.17 & 36.75±1.47 \\
Chemical &    RAR &   0.7 & TS-PubMedBERT-Partial-CRF &  81.00±1.11 &  90.04±1.57 & 85.26±0.40 \\
Chemical &    RAR &   0.8 &        BiLSTM-Partial-CRF &  78.85±3.34 &   9.57±0.70 & 17.06±1.17 \\
Chemical &    RAR &   0.8 &            EER-PubMedBERT &  84.46±0.95 &  87.01±1.37 & 85.70±0.25 \\
Chemical &    RAR &   0.8 &                PubMedBERT &  87.55±0.64 &  12.68±1.58 & 22.11±2.37 \\
Chemical &    RAR &   0.8 & TS-PubMedBERT-Partial-CRF &  83.42±0.69 &  88.97±0.57 & 86.10±0.22 \\
Chemical &    RAR &   0.9 &        BiLSTM-Partial-CRF &  65.16±2.69 &   2.76±0.60 &  5.29±1.10 \\
Chemical &    RAR &   0.9 &            EER-PubMedBERT &  84.85±0.45 &  84.30±0.85 & 84.57±0.21 \\
Chemical &    RAR &   0.9 &                PubMedBERT &  84.96±1.72 &   5.25±0.37 &  9.89±0.67 \\
Chemical &    RAR &   0.9 & TS-PubMedBERT-Partial-CRF &  84.21±0.85 &  86.07±0.89 & 85.12±0.55 \\
\hline
Chemical &   RSFR &   0.1 &        BiLSTM-Partial-CRF &  79.02±0.56 &  78.09±0.64 & 78.55±0.33 \\
Chemical &   RSFR &   0.1 &            EER-PubMedBERT &  82.28±0.52 &  86.37±0.42 & 84.27±0.26 \\
Chemical &   RSFR &   0.1 &                PubMedBERT &  86.58±0.23 &  79.26±0.44 & 82.76±0.31 \\
Chemical &   RSFR &   0.1 & TS-PubMedBERT-Partial-CRF &  74.46±2.59 &  90.70±0.94 & 81.76±1.70 \\
Chemical &   RSFR &   0.2 &        BiLSTM-Partial-CRF &  77.71±0.42 &  70.14±0.46 & 73.73±0.28 \\
Chemical &   RSFR &   0.2 &            EER-PubMedBERT &  79.10±2.55 &  84.71±2.90 & 81.72±0.67 \\
Chemical &   RSFR &   0.2 &                PubMedBERT &  86.37±0.25 &  70.68±0.30 & 77.74±0.19 \\
Chemical &   RSFR &   0.2 & TS-PubMedBERT-Partial-CRF &  74.62±2.89 &  90.38±0.29 & 81.72±1.77 \\
Chemical &   RSFR &   0.3 &        BiLSTM-Partial-CRF &  76.01±0.32 &  64.02±1.32 & 69.49±0.71 \\
Chemical &   RSFR &   0.3 &            EER-PubMedBERT &  74.81±1.12 &  84.78±1.83 & 79.46±0.46 \\
Chemical &   RSFR &   0.3 &                PubMedBERT &  84.93±1.05 &  62.39±0.98 & 71.92±0.44 \\
Chemical &   RSFR &   0.3 & TS-PubMedBERT-Partial-CRF &  74.00±2.33 &  88.73±0.90 & 80.67±1.21 \\
Chemical &   RSFR &   0.4 &        BiLSTM-Partial-CRF &  75.01±1.08 &  58.42±0.91 & 65.67±0.26 \\
Chemical &   RSFR &   0.4 &            EER-PubMedBERT &  75.27±1.92 &  79.29±2.13 & 77.18±0.44 \\
Chemical &   RSFR &   0.4 &                PubMedBERT &  82.22±1.92 &  57.94±2.30 & 67.93±1.58 \\
Chemical &   RSFR &   0.4 & TS-PubMedBERT-Partial-CRF &  74.15±0.72 &  85.84±1.10 & 79.56±0.62 \\
Chemical &   RSFR &   0.5 &        BiLSTM-Partial-CRF &  75.49±0.73 &  50.19±1.00 & 60.28±0.49 \\
Chemical &   RSFR &   0.5 &            EER-PubMedBERT &  76.85±1.29 &  71.56±0.64 & 74.10±0.35 \\
Chemical &   RSFR &   0.5 &                PubMedBERT &  82.61±1.41 &  49.76±1.05 & 62.11±1.10 \\
Chemical &   RSFR &   0.5 & TS-PubMedBERT-Partial-CRF &  73.23±2.29 &  81.50±1.19 & 77.11±1.23 \\
Chemical &   RSFR &   0.6 &        BiLSTM-Partial-CRF &  74.99±0.75 &  42.14±0.62 & 53.96±0.42 \\
Chemical &   RSFR &   0.6 &            EER-PubMedBERT &  77.12±0.66 &  64.64±1.34 & 70.32±0.96 \\
Chemical &   RSFR &   0.6 &                PubMedBERT &  82.38±1.20 &  38.11±1.66 & 52.08±1.35 \\
Chemical &   RSFR &   0.6 & TS-PubMedBERT-Partial-CRF &  72.62±3.65 &  76.28±3.28 & 74.24±0.99 \\
Chemical &   RSFR &   0.7 &        BiLSTM-Partial-CRF &  74.98±1.32 &  34.84±1.33 & 47.55±1.26 \\
Chemical &   RSFR &   0.7 &            EER-PubMedBERT &  77.57±1.57 &  54.80±2.87 & 64.15±1.53 \\
Chemical &   RSFR &   0.7 &                PubMedBERT &  82.13±1.59 &  30.29±2.97 & 44.16±3.22 \\
Chemical &   RSFR &   0.7 & TS-PubMedBERT-Partial-CRF &  70.35±3.82 &  72.82±3.51 & 71.41±1.61 \\
Chemical &   RSFR &   0.8 &        BiLSTM-Partial-CRF &  76.64±0.54 &  25.43±0.24 & 38.19±0.28 \\
Chemical &   RSFR &   0.8 &            EER-PubMedBERT &  77.21±2.53 &  45.08±1.18 & 56.90±1.05 \\
Chemical &   RSFR &   0.8 &                PubMedBERT &  77.16±1.82 &  19.09±1.08 & 30.60±1.49 \\
Chemical &   RSFR &   0.8 & TS-PubMedBERT-Partial-CRF &  74.37±4.58 &  62.55±4.15 & 67.72±1.84 \\
Chemical &   RSFR &   0.9 &        BiLSTM-Partial-CRF &  75.85±1.63 &  13.81±0.68 & 23.35±0.93 \\
Chemical &   RSFR &   0.9 &            EER-PubMedBERT &  77.61±1.38 &  35.46±1.44 & 48.66±1.45 \\
Chemical &   RSFR &   0.9 &                PubMedBERT &  70.53±7.73 &   8.83±1.78 & 15.64±2.90 \\
Chemical &   RSFR &   0.9 & TS-PubMedBERT-Partial-CRF &  73.89±1.53 &  46.85±1.84 & 57.30±1.02 \\
\hline
    \hline
 Disease &    RAR &   0.1 &        BiLSTM-Partial-CRF &  77.19±0.37 &  82.62±0.56 & 79.81±0.12 \\
 Disease &    RAR &   0.1 &            EER-PubMedBERT &  80.87±1.20 &  87.81±0.59 & 84.19±0.44 \\
 Disease &    RAR &   0.1 &                PubMedBERT &  83.32±0.75 &  84.54±1.79 & 83.91±0.54 \\
 Disease &    RAR &   0.1 & TS-PubMedBERT-Partial-CRF &  79.51±1.11 &  88.48±0.65 & 83.75±0.35 \\
 Disease &    RAR &   0.2 &        BiLSTM-Partial-CRF &  77.50±0.59 &  81.87±0.78 & 79.62±0.23 \\
 Disease &    RAR &   0.2 &            EER-PubMedBERT &  81.31±1.37 &  87.36±0.41 & 84.22±0.71 \\
 Disease &    RAR &   0.2 &                PubMedBERT &  83.50±0.84 &  83.08±1.53 & 83.27±0.38 \\
 Disease &    RAR &   0.2 & TS-PubMedBERT-Partial-CRF &  79.39±0.85 &  88.31±0.20 & 83.61±0.45 \\
 Disease &    RAR &   0.3 &        BiLSTM-Partial-CRF &  77.21±0.46 &  81.48±0.28 & 79.28±0.27 \\
 Disease &    RAR &   0.3 &            EER-PubMedBERT &  82.67±0.93 &  85.89±0.48 & 84.24±0.26 \\
 Disease &    RAR &   0.3 &                PubMedBERT &  82.39±1.34 &  81.42±1.04 & 81.89±0.57 \\
 Disease &    RAR &   0.3 & TS-PubMedBERT-Partial-CRF &  79.94±0.81 &  88.35±0.29 & 83.93±0.47 \\
 Disease &    RAR &   0.4 &        BiLSTM-Partial-CRF &  77.44±0.43 &  80.75±0.43 & 79.06±0.26 \\
 Disease &    RAR &   0.4 &            EER-PubMedBERT &  82.53±0.75 &  85.30±0.56 & 83.89±0.28 \\
 Disease &    RAR &   0.4 &                PubMedBERT &  82.58±1.99 &  76.94±3.24 & 79.64±2.55 \\
 Disease &    RAR &   0.4 & TS-PubMedBERT-Partial-CRF &  79.98±1.34 &  87.78±0.36 & 83.69±0.61 \\
 Disease &    RAR &   0.5 &        BiLSTM-Partial-CRF &  78.19±0.52 &  78.93±0.49 & 78.55±0.27 \\
 Disease &    RAR &   0.5 &            EER-PubMedBERT &  82.01±1.52 &  84.70±1.10 & 83.31±0.27 \\
 Disease &    RAR &   0.5 &                PubMedBERT &  85.86±1.56 &  58.51±5.58 & 69.40±3.52 \\
 Disease &    RAR &   0.5 & TS-PubMedBERT-Partial-CRF &  80.36±1.31 &  87.30±0.43 & 83.68±0.61 \\
 Disease &    RAR &   0.6 &        BiLSTM-Partial-CRF &  78.54±0.86 &  76.88±0.85 & 77.69±0.25 \\
 Disease &    RAR &   0.6 &            EER-PubMedBERT &  81.74±1.12 &  83.83±1.01 & 82.76±0.24 \\
 Disease &    RAR &   0.6 &                PubMedBERT &  85.68±1.62 &  45.03±6.91 & 58.69±5.80 \\
 Disease &    RAR &   0.6 & TS-PubMedBERT-Partial-CRF &  80.28±1.09 &  86.39±1.00 & 83.21±0.28 \\
 Disease &    RAR &   0.7 &        BiLSTM-Partial-CRF &  79.20±0.67 &  70.18±1.98 & 74.40±1.19 \\
 Disease &    RAR &   0.7 &            EER-PubMedBERT &  81.35±1.10 &  82.22±1.01 & 81.77±0.56 \\
 Disease &    RAR &   0.7 &                PubMedBERT &  85.16±0.45 &  26.40±1.68 & 40.27±1.96 \\
 Disease &    RAR &   0.7 & TS-PubMedBERT-Partial-CRF &  80.72±1.27 &  85.18±0.48 & 82.88±0.66 \\
 Disease &    RAR &   0.8 &        BiLSTM-Partial-CRF &  73.55±4.45 &  37.96±9.51 & 49.55±8.71 \\
 Disease &    RAR &   0.8 &            EER-PubMedBERT &  80.32±0.82 &  80.94±1.51 & 80.61±0.39 \\
 Disease &    RAR &   0.8 &                PubMedBERT &  82.40±1.90 &  15.28±2.13 & 25.71±3.03 \\
 Disease &    RAR &   0.8 & TS-PubMedBERT-Partial-CRF &  80.38±1.70 &  83.19±1.06 & 81.73±0.42 \\
 Disease &    RAR &   0.9 &        BiLSTM-Partial-CRF &  71.04±8.76 &   9.29±6.82 & 15.50±9.36 \\
 Disease &    RAR &   0.9 &            EER-PubMedBERT &  79.13±0.85 &  78.09±0.75 & 78.60±0.43 \\
 Disease &    RAR &   0.9 &                PubMedBERT &  82.07±4.90 &   5.19±1.33 &  9.73±2.36 \\
 Disease &    RAR &   0.9 & TS-PubMedBERT-Partial-CRF &  79.32±1.60 &  81.41±1.55 & 80.32±0.64 \\
 \hline
 Disease &   RSFR &   0.1 &        BiLSTM-Partial-CRF &  75.91±0.50 &  76.15±0.81 & 76.03±0.46 \\
 Disease &   RSFR &   0.1 &            EER-PubMedBERT &  81.61±1.10 &  85.43±1.25 & 83.46±0.43 \\
 Disease &   RSFR &   0.1 &                PubMedBERT &  83.99±1.07 &  77.54±1.15 & 80.62±0.49 \\
 Disease &   RSFR &   0.1 & TS-PubMedBERT-Partial-CRF &  79.44±0.58 &  87.07±0.81 & 83.08±0.40 \\
 Disease &   RSFR &   0.2 &        BiLSTM-Partial-CRF &  75.44±0.62 &  70.07±1.32 & 72.65±0.58 \\
 Disease &   RSFR &   0.2 &            EER-PubMedBERT &  79.57±1.43 &  84.22±1.53 & 81.80±0.28 \\
 Disease &   RSFR &   0.2 &                PubMedBERT &  82.31±0.80 &  69.39±0.52 & 75.29±0.51 \\
 Disease &   RSFR &   0.2 & TS-PubMedBERT-Partial-CRF &  79.08±0.63 &  86.28±1.29 & 82.51±0.53 \\
 Disease &   RSFR &   0.3 &        BiLSTM-Partial-CRF &  73.19±0.72 &  63.97±0.71 & 68.27±0.42 \\
 Disease &   RSFR &   0.3 &            EER-PubMedBERT &  77.25±0.57 &  82.83±0.64 & 79.94±0.37 \\
 Disease &   RSFR &   0.3 &                PubMedBERT &  80.54±1.95 &  61.22±0.85 & 69.56±1.10 \\
 Disease &   RSFR &   0.3 & TS-PubMedBERT-Partial-CRF &  79.02±0.96 &  83.50±0.70 & 81.20±0.75 \\
 Disease &   RSFR &   0.4 &        BiLSTM-Partial-CRF &  71.83±0.79 &  58.94±1.38 & 64.73±0.62 \\
 Disease &   RSFR &   0.4 &            EER-PubMedBERT &  75.79±1.02 &  80.53±0.97 & 78.08±0.44 \\
 Disease &   RSFR &   0.4 &                PubMedBERT &  79.21±0.99 &  51.62±1.11 & 62.49±0.63 \\
 Disease &   RSFR &   0.4 & TS-PubMedBERT-Partial-CRF &  79.49±1.18 &  83.13±0.67 & 81.26±0.59 \\
 Disease &   RSFR &   0.5 &        BiLSTM-Partial-CRF &  68.23±1.08 &  53.84±0.32 & 60.18±0.47 \\
 Disease &   RSFR &   0.5 &            EER-PubMedBERT &  74.80±1.58 &  78.37±1.84 & 76.51±0.57 \\
 Disease &   RSFR &   0.5 &                PubMedBERT &  78.19±1.00 &  43.21±0.34 & 55.65±0.35 \\
 Disease &   RSFR &   0.5 & TS-PubMedBERT-Partial-CRF &  80.14±0.86 &  80.18±0.82 & 80.16±0.81 \\
 Disease &   RSFR &   0.6 &        BiLSTM-Partial-CRF &  67.50±0.65 &  48.32±0.39 & 56.32±0.34 \\
 Disease &   RSFR &   0.6 &            EER-PubMedBERT &  73.96±1.16 &  68.23±2.49 & 70.93±1.05 \\
 Disease &   RSFR &   0.6 &                PubMedBERT &  75.11±2.37 &  34.38±1.30 & 47.13±1.19 \\
 Disease &   RSFR &   0.6 & TS-PubMedBERT-Partial-CRF &  78.60±1.49 &  77.10±1.78 & 77.81±0.30 \\
 Disease &   RSFR &   0.7 &        BiLSTM-Partial-CRF &  65.33±1.77 &  40.54±0.75 & 50.03±1.06 \\
 Disease &   RSFR &   0.7 &            EER-PubMedBERT &  71.10±1.33 &  58.98±3.43 & 64.40±1.82 \\
 Disease &   RSFR &   0.7 &                PubMedBERT &  73.14±2.26 &  27.05±2.06 & 39.45±2.30 \\
 Disease &   RSFR &   0.7 & TS-PubMedBERT-Partial-CRF &  77.88±1.40 &  72.00±1.41 & 74.81±1.15 \\
 Disease &   RSFR &   0.8 &        BiLSTM-Partial-CRF &  63.93±2.80 &  31.13±1.80 & 41.77±0.93 \\
 Disease &   RSFR &   0.8 &            EER-PubMedBERT &  70.78±1.86 &  54.49±4.10 & 61.45±2.26 \\
 Disease &   RSFR &   0.8 &                PubMedBERT &  67.93±2.45 &  17.36±1.24 & 27.62±1.52 \\
 Disease &   RSFR &   0.8 & TS-PubMedBERT-Partial-CRF &  77.58±1.39 &  71.18±3.85 & 74.18±2.11 \\
 Disease &   RSFR &   0.9 &        BiLSTM-Partial-CRF &  58.21±4.42 &  18.51±5.61 & 27.46±5.89 \\
 Disease &   RSFR &   0.9 &            EER-PubMedBERT &  70.67±1.51 &  45.43±3.96 & 55.19±2.89 \\
 Disease &   RSFR &   0.9 &                PubMedBERT &  61.52±4.37 &   7.07±1.21 & 12.64±1.92 \\
 Disease &   RSFR &   0.9 & TS-PubMedBERT-Partial-CRF &  76.12±2.52 &  67.92±5.52 & 71.71±3.94 \\
 \hline
     \hline
    Gene &    RAR &   0.1 &        BiLSTM-Partial-CRF &  69.66±0.33 &  76.97±0.42 & 73.13±0.12 \\
    Gene &    RAR &   0.1 &            EER-PubMedBERT &  71.22±0.23 &  83.48±0.31 & 76.86±0.23 \\
    Gene &    RAR &   0.1 &                PubMedBERT &  75.22±0.47 &  79.05±0.64 & 77.08±0.16 \\
    Gene &    RAR &   0.1 & TS-PubMedBERT-Partial-CRF &  59.81±1.06 &  78.24±0.90 & 67.79±0.86 \\
    Gene &    RAR &   0.2 &        BiLSTM-Partial-CRF &  70.12±0.44 &  76.41±0.38 & 73.13±0.16 \\
    Gene &    RAR &   0.2 &            EER-PubMedBERT &  71.60±0.42 &  83.10±0.17 & 76.92±0.29 \\
    Gene &    RAR &   0.2 &                PubMedBERT &  76.67±0.90 &  74.37±1.83 & 75.48±0.56 \\
    Gene &    RAR &   0.2 & TS-PubMedBERT-Partial-CRF &  59.83±0.83 &  78.03±0.49 & 67.73±0.68 \\
    Gene &    RAR &   0.3 &        BiLSTM-Partial-CRF &  71.64±0.79 &  74.17±0.77 & 72.87±0.14 \\
    Gene &    RAR &   0.3 &            EER-PubMedBERT &  72.54±0.38 &  82.65±0.22 & 77.26±0.14 \\
    Gene &    RAR &   0.3 &                PubMedBERT &  76.13±0.59 &  73.05±0.63 & 74.56±0.44 \\
    Gene &    RAR &   0.3 & TS-PubMedBERT-Partial-CRF &  60.49±1.22 &  78.39±1.54 & 68.28±1.24 \\
    Gene &    RAR &   0.4 &        BiLSTM-Partial-CRF &  72.76±0.74 &  71.10±0.59 & 71.92±0.14 \\
    Gene &    RAR &   0.4 &            EER-PubMedBERT &  73.44±0.44 &  81.82±0.22 & 77.41±0.19 \\
    Gene &    RAR &   0.4 &                PubMedBERT &  77.50±0.47 &  66.93±1.64 & 71.81±0.76 \\
    Gene &    RAR &   0.4 & TS-PubMedBERT-Partial-CRF &  63.36±1.29 &  79.51±0.59 & 70.52±0.96 \\
    Gene &    RAR &   0.5 &        BiLSTM-Partial-CRF &  75.47±0.59 &  63.00±0.40 & 68.67±0.22 \\
    Gene &    RAR &   0.5 &            EER-PubMedBERT &  74.39±0.15 &  80.49±0.27 & 77.32±0.06 \\
    Gene &    RAR &   0.5 &                PubMedBERT &  78.02±1.00 &  45.55±6.39 & 57.22±4.87 \\
    Gene &    RAR &   0.5 & TS-PubMedBERT-Partial-CRF &  62.69±2.70 &  78.54±1.79 & 69.71±2.25 \\
    Gene &    RAR &   0.6 &        BiLSTM-Partial-CRF &  78.61±1.13 &  46.12±2.73 & 58.06±1.98 \\
    Gene &    RAR &   0.6 &            EER-PubMedBERT &  74.48±1.18 &  79.42±1.73 & 76.84±0.35 \\
    Gene &    RAR &   0.6 &                PubMedBERT &  80.00±0.51 &  29.11±2.13 & 42.65±2.23 \\
    Gene &    RAR &   0.6 & TS-PubMedBERT-Partial-CRF &  65.56±1.88 &  80.64±1.09 & 72.31±1.52 \\
    Gene &    RAR &   0.7 &        BiLSTM-Partial-CRF &  73.79±3.63 &  16.29±1.28 & 26.67±1.90 \\
    Gene &    RAR &   0.7 &            EER-PubMedBERT &  75.14±0.53 &  77.54±0.92 & 76.32±0.22 \\
    Gene &    RAR &   0.7 &                PubMedBERT &  80.06±0.83 &  18.95±1.58 & 30.61±2.12 \\
    Gene &    RAR &   0.7 & TS-PubMedBERT-Partial-CRF &  68.35±0.73 &  81.45±0.83 & 74.32±0.57 \\
    Gene &    RAR &   0.8 &        BiLSTM-Partial-CRF &  74.76±3.01 &   4.46±1.40 &  8.37±2.48 \\
    Gene &    RAR &   0.8 &            EER-PubMedBERT &  74.69±1.31 &  76.24±2.09 & 75.42±0.51 \\
    Gene &    RAR &   0.8 &                PubMedBERT &  78.17±2.32 &   9.66±0.76 & 17.19±1.22 \\
    Gene &    RAR &   0.8 & TS-PubMedBERT-Partial-CRF &  69.15±0.82 &  81.70±0.78 & 74.90±0.62 \\
    Gene &    RAR &   0.9 &        BiLSTM-Partial-CRF & 50.85±14.60 &   0.18±0.13 &  0.36±0.25 \\
    Gene &    RAR &   0.9 &            EER-PubMedBERT &  76.99±0.78 &  71.18±1.16 & 73.96±0.30 \\
    Gene &    RAR &   0.9 &                PubMedBERT &  76.55±2.71 &   3.52±0.23 &  6.72±0.43 \\
    Gene &    RAR &   0.9 & TS-PubMedBERT-Partial-CRF &  73.52±1.22 &  77.76±1.04 & 75.56±0.35 \\
    \hline
    Gene &   RSFR &   0.1 &        BiLSTM-Partial-CRF &  67.13±0.32 &  69.92±0.66 & 68.49±0.24 \\
    Gene &   RSFR &   0.1 &            EER-PubMedBERT &  70.36±0.86 &  77.95±0.57 & 73.96±0.71 \\
    Gene &   RSFR &   0.1 &                PubMedBERT &  75.16±0.49 &  72.29±0.97 & 73.69±0.31 \\
    Gene &   RSFR &   0.1 & TS-PubMedBERT-Partial-CRF &  58.47±0.56 &  77.31±0.62 & 66.59±0.57 \\
    Gene &   RSFR &   0.2 &        BiLSTM-Partial-CRF &  65.96±0.25 &  63.21±0.50 & 64.56±0.19 \\
    Gene &   RSFR &   0.2 &            EER-PubMedBERT &  70.23±0.31 &  72.58±0.66 & 71.38±0.30 \\
    Gene &   RSFR &   0.2 &                PubMedBERT &  71.98±0.84 &  66.67±1.07 & 69.22±0.74 \\
    Gene &   RSFR &   0.2 & TS-PubMedBERT-Partial-CRF &  58.30±1.41 &  75.87±1.54 & 65.93±1.45 \\
    Gene &   RSFR &   0.3 &        BiLSTM-Partial-CRF &  64.59±0.40 &  56.63±0.56 & 60.35±0.30 \\
    Gene &   RSFR &   0.3 &            EER-PubMedBERT &  68.07±1.57 &  68.80±1.89 & 68.39±0.23 \\
    Gene &   RSFR &   0.3 &                PubMedBERT &  72.58±1.09 &  57.95±0.92 & 64.43±0.29 \\
    Gene &   RSFR &   0.3 & TS-PubMedBERT-Partial-CRF &  57.65±1.83 &  74.23±0.88 & 64.88±1.34 \\
    Gene &   RSFR &   0.4 &        BiLSTM-Partial-CRF &  62.01±0.48 &  50.99±0.82 & 55.95±0.38 \\
    Gene &   RSFR &   0.4 &            EER-PubMedBERT &  66.09±1.48 &  65.70±1.39 & 65.87±0.45 \\
    Gene &   RSFR &   0.4 &                PubMedBERT &  71.20±2.01 &  50.85±2.82 & 59.25±1.60 \\
    Gene &   RSFR &   0.4 & TS-PubMedBERT-Partial-CRF &  58.77±2.30 &  73.76±1.26 & 65.40±1.68 \\
    Gene &   RSFR &   0.5 &        BiLSTM-Partial-CRF &  61.89±1.02 &  43.72±0.54 & 51.23±0.26 \\
    Gene &   RSFR &   0.5 &            EER-PubMedBERT &  63.70±3.10 &  62.38±2.39 & 62.91±0.52 \\
    Gene &   RSFR &   0.5 &                PubMedBERT &  70.70±2.54 &  41.86±1.61 & 52.52±0.91 \\
    Gene &   RSFR &   0.5 & TS-PubMedBERT-Partial-CRF &  58.44±0.96 &  69.47±2.01 & 63.47±1.18 \\
    Gene &   RSFR &   0.6 &        BiLSTM-Partial-CRF &  59.70±0.81 &  36.23±0.92 & 45.08±0.50 \\
    Gene &   RSFR &   0.6 &            EER-PubMedBERT &  63.78±2.28 &  54.88±2.23 & 58.91±0.54 \\
    Gene &   RSFR &   0.6 &                PubMedBERT &  69.76±1.97 &  35.32±2.43 & 46.86±2.45 \\
    Gene &   RSFR &   0.6 & TS-PubMedBERT-Partial-CRF &  57.52±1.85 &  65.80±1.73 & 61.36±1.31 \\
    Gene &   RSFR &   0.7 &        BiLSTM-Partial-CRF &  59.34±1.71 &  25.22±1.13 & 35.36±0.88 \\
    Gene &   RSFR &   0.7 &            EER-PubMedBERT &  60.96±2.24 &  49.71±2.08 & 54.73±1.68 \\
    Gene &   RSFR &   0.7 &                PubMedBERT &  66.75±1.74 &  24.39±3.41 & 35.62±3.93 \\
    Gene &   RSFR &   0.7 & TS-PubMedBERT-Partial-CRF &  56.37±2.54 &  63.49±2.89 & 59.60±0.66 \\
    Gene &   RSFR &   0.8 &        BiLSTM-Partial-CRF &  58.71±1.36 &  14.47±0.86 & 23.19±1.06 \\
    Gene &   RSFR &   0.8 &            EER-PubMedBERT &  59.55±1.87 &  42.34±1.56 & 49.44±0.76 \\
    Gene &   RSFR &   0.8 &                PubMedBERT &  64.97±1.97 &  13.76±1.84 & 22.65±2.52 \\
    Gene &   RSFR &   0.8 & TS-PubMedBERT-Partial-CRF &  55.44±2.39 &  58.74±5.04 & 56.81±1.34 \\
    Gene &   RSFR &   0.9 &        BiLSTM-Partial-CRF &  52.14±6.95 &   4.08±0.69 &  7.53±1.12 \\
    Gene &   RSFR &   0.9 &            EER-PubMedBERT &  58.08±1.00 &  33.03±3.13 & 42.00±2.52 \\
    Gene &   RSFR &   0.9 &                PubMedBERT &  62.60±3.09 &   4.48±0.93 &  8.33±1.61 \\
    Gene &   RSFR &   0.9 & TS-PubMedBERT-Partial-CRF &  56.08±2.59 &  49.94±2.42 & 52.79±2.05 \\
    \hline
        \hline
 Species &    RAR &   0.1 &        BiLSTM-Partial-CRF &  73.45±0.74 &  71.96±0.85 & 72.69±0.62 \\
 Species &    RAR &   0.1 &            EER-PubMedBERT &  55.67±2.80 &  85.64±0.41 & 67.43±2.01 \\
 Species &    RAR &   0.1 &                PubMedBERT &  77.69±1.15 &  74.87±1.97 & 76.22±0.71 \\
 Species &    RAR &   0.1 & TS-PubMedBERT-Partial-CRF &  56.01±1.04 &  85.13±0.78 & 67.56±0.82 \\
 Species &    RAR &   0.2 &        BiLSTM-Partial-CRF &  74.25±0.67 &  70.67±0.21 & 72.42±0.35 \\
 Species &    RAR &   0.2 &            EER-PubMedBERT &  55.35±1.42 &  86.30±0.46 & 67.43±1.03 \\
 Species &    RAR &   0.2 &                PubMedBERT &  78.19±1.40 &  73.60±1.43 & 75.80±0.57 \\
 Species &    RAR &   0.2 & TS-PubMedBERT-Partial-CRF &  57.83±1.95 &  85.23±0.60 & 68.88±1.33 \\
 Species &    RAR &   0.3 &        BiLSTM-Partial-CRF &  75.16±1.16 &  69.26±0.43 & 72.08±0.45 \\
 Species &    RAR &   0.3 &            EER-PubMedBERT &  56.55±1.80 &  85.56±0.54 & 68.07±1.19 \\
 Species &    RAR &   0.3 &                PubMedBERT &  79.10±1.97 &  66.96±3.34 & 72.43±1.28 \\
 Species &    RAR &   0.3 & TS-PubMedBERT-Partial-CRF &  56.99±2.81 &  85.97±0.59 & 68.49±1.94 \\
 Species &    RAR &   0.4 &        BiLSTM-Partial-CRF &  76.04±1.86 &  68.27±0.83 & 71.93±0.55 \\
 Species &    RAR &   0.4 &            EER-PubMedBERT &  55.70±3.13 &  85.63±0.57 & 67.43±2.16 \\
 Species &    RAR &   0.4 &                PubMedBERT &  79.59±1.83 &  65.79±4.52 & 71.89±2.25 \\
 Species &    RAR &   0.4 & TS-PubMedBERT-Partial-CRF &  57.03±2.80 &  85.65±1.26 & 68.40±1.65 \\
 Species &    RAR &   0.5 &        BiLSTM-Partial-CRF &  76.48±1.19 &  66.32±0.89 & 71.03±0.84 \\
 Species &    RAR &   0.5 &            EER-PubMedBERT &  58.27±1.47 &  85.09±0.75 & 69.16±0.98 \\
 Species &    RAR &   0.5 &                PubMedBERT &  79.26±0.94 &  57.61±7.44 & 66.40±4.97 \\
 Species &    RAR &   0.5 & TS-PubMedBERT-Partial-CRF &  57.91±1.88 &  85.30±0.60 & 68.96±1.17 \\
 Species &    RAR &   0.6 &        BiLSTM-Partial-CRF &  84.19±0.38 &  53.16±0.88 & 65.17±0.54 \\
 Species &    RAR &   0.6 &            EER-PubMedBERT &  57.85±1.36 &  84.99±0.24 & 68.83±0.96 \\
 Species &    RAR &   0.6 &                PubMedBERT &  77.64±2.34 &  41.73±4.14 & 54.15±3.58 \\
 Species &    RAR &   0.6 & TS-PubMedBERT-Partial-CRF &  55.79±2.18 &  85.33±0.48 & 67.44±1.48 \\
 Species &    RAR &   0.7 &        BiLSTM-Partial-CRF &  77.22±1.56 &  59.29±1.08 & 67.06±0.49 \\
 Species &    RAR &   0.7 &            EER-PubMedBERT &  56.86±2.89 &  85.05±0.92 & 68.09±1.83 \\
 Species &    RAR &   0.7 &                PubMedBERT &  77.75±2.76 &  25.17±2.66 & 37.95±3.08 \\
 Species &    RAR &   0.7 & TS-PubMedBERT-Partial-CRF &  53.87±2.74 &  85.45±0.68 & 66.03±1.98 \\
 Species &    RAR &   0.8 &        BiLSTM-Partial-CRF &  79.31±2.39 & 45.40±10.71 & 56.81±9.81 \\
 Species &    RAR &   0.8 &            EER-PubMedBERT &  58.18±2.32 &  82.97±0.60 & 68.37±1.55 \\
 Species &    RAR &   0.8 &                PubMedBERT &  71.62±2.65 &  16.16±3.95 & 26.10±5.35 \\
 Species &    RAR &   0.8 & TS-PubMedBERT-Partial-CRF &  57.78±5.39 &  82.24±2.72 & 67.62±2.81 \\
 Species &    RAR &   0.9 &        BiLSTM-Partial-CRF &  78.96±8.18 &   4.74±3.51 &  8.72±6.18 \\
 Species &    RAR &   0.9 &            EER-PubMedBERT &  59.94±2.41 &  78.96±4.71 & 68.01±1.76 \\
 Species &    RAR &   0.9 &                PubMedBERT &  67.87±5.36 &   5.20±2.59 &  9.46±4.20 \\
 Species &    RAR &   0.9 & TS-PubMedBERT-Partial-CRF &  54.72±5.83 &  82.40±1.14 & 65.53±4.31 \\
 \hline
 Species &   RSFR &   0.1 &        BiLSTM-Partial-CRF &  70.34±2.21 &  58.33±3.62 & 63.75±2.99 \\
 Species &   RSFR &   0.1 &            EER-PubMedBERT &  55.66±1.26 &  85.59±0.32 & 67.44±0.84 \\
 Species &   RSFR &   0.1 &                PubMedBERT &  75.18±1.19 &  62.98±1.82 & 68.53±1.39 \\
 Species &   RSFR &   0.1 & TS-PubMedBERT-Partial-CRF &  55.85±1.76 &  85.76±0.67 & 67.62±1.11 \\
 Species &   RSFR &   0.2 &        BiLSTM-Partial-CRF &  67.98±1.63 &  52.81±2.28 & 59.41±1.66 \\
 Species &   RSFR &   0.2 &            EER-PubMedBERT &  55.79±1.29 &  85.95±0.30 & 67.66±1.01 \\
 Species &   RSFR &   0.2 &                PubMedBERT &  74.57±1.72 &  54.87±3.42 & 63.15±2.40 \\
 Species &   RSFR &   0.2 & TS-PubMedBERT-Partial-CRF &  55.42±2.94 &  85.14±1.03 & 67.08±2.04 \\
 Species &   RSFR &   0.3 &        BiLSTM-Partial-CRF &  66.37±1.36 &  46.66±1.68 & 54.77±1.30 \\
 Species &   RSFR &   0.3 &            EER-PubMedBERT &  56.61±2.23 &  81.55±2.36 & 66.76±1.20 \\
 Species &   RSFR &   0.3 &                PubMedBERT &  71.79±2.53 &  49.93±1.27 & 58.88±1.51 \\
 Species &   RSFR &   0.3 & TS-PubMedBERT-Partial-CRF &  56.38±1.82 &  83.36±2.16 & 67.24±1.53 \\
 Species &   RSFR &   0.4 &        BiLSTM-Partial-CRF &  66.13±1.50 &  40.93±0.85 & 50.56±0.93 \\
 Species &   RSFR &   0.4 &            EER-PubMedBERT &  56.66±1.32 &  78.84±3.35 & 65.89±1.50 \\
 Species &   RSFR &   0.4 &                PubMedBERT &  71.94±2.56 &  43.43±2.17 & 54.08±1.25 \\
 Species &   RSFR &   0.4 & TS-PubMedBERT-Partial-CRF &  57.34±3.54 &  82.03±2.43 & 67.38±2.19 \\
 Species &   RSFR &   0.5 &        BiLSTM-Partial-CRF &  66.75±2.52 &  36.66±1.45 & 47.29±1.27 \\
 Species &   RSFR &   0.5 &            EER-PubMedBERT &  53.89±1.58 &  76.57±5.06 & 63.21±2.56 \\
 Species &   RSFR &   0.5 &                PubMedBERT &  69.93±1.97 &  39.10±1.71 & 50.10±1.02 \\
 Species &   RSFR &   0.5 & TS-PubMedBERT-Partial-CRF &  56.52±1.75 &  79.69±2.98 & 66.12±1.97 \\
 Species &   RSFR &   0.6 &        BiLSTM-Partial-CRF &  65.83±1.74 &  32.20±1.17 & 43.21±0.78 \\
 Species &   RSFR &   0.6 &            EER-PubMedBERT &  53.71±1.44 &  71.20±6.12 & 61.04±1.34 \\
 Species &   RSFR &   0.6 &                PubMedBERT &  69.93±2.00 &  34.03±1.76 & 45.73±1.32 \\
 Species &   RSFR &   0.6 & TS-PubMedBERT-Partial-CRF &  55.18±1.77 &  81.33±3.44 & 65.74±2.31 \\
 Species &   RSFR &   0.7 &        BiLSTM-Partial-CRF &  65.95±2.09 &  27.47±0.92 & 38.76±0.79 \\
 Species &   RSFR &   0.7 &            EER-PubMedBERT &  54.18±0.71 &  66.57±3.27 & 59.71±1.67 \\
 Species &   RSFR &   0.7 &                PubMedBERT &  71.19±1.45 &  26.16±1.94 & 38.21±2.11 \\
 Species &   RSFR &   0.7 & TS-PubMedBERT-Partial-CRF &  53.99±4.10 &  77.92±5.81 & 63.50±2.30 \\
 Species &   RSFR &   0.8 &        BiLSTM-Partial-CRF &  66.51±3.37 &  22.56±1.10 & 33.65±1.26 \\
 Species &   RSFR &   0.8 &            EER-PubMedBERT &  53.77±2.73 &  60.94±3.52 & 56.99±1.37 \\
 Species &   RSFR &   0.8 &                PubMedBERT &  65.73±2.86 &  18.54±1.68 & 28.84±1.93 \\
 Species &   RSFR &   0.8 & TS-PubMedBERT-Partial-CRF &  57.67±4.05 &  71.64±9.56 & 63.20±2.19 \\
 Species &   RSFR &   0.9 &        BiLSTM-Partial-CRF &  68.68±4.78 &  12.88±2.70 & 21.52±3.71 \\
 Species &   RSFR &   0.9 &            EER-PubMedBERT &  49.28±3.44 &  60.60±1.66 & 54.24±1.52 \\
 Species &   RSFR &   0.9 &                PubMedBERT &  61.03±4.84 &   8.73±2.43 & 15.09±3.44 \\
 Species &   RSFR &   0.9 & TS-PubMedBERT-Partial-CRF &  50.58±5.85 &  65.81±4.18 & 56.88±3.53 \\
\end{supertabular}
\end{center}
\end{appendices}
\end{document}